\pdfoutput=1

\documentclass[11pt]{article}

\usepackage[]{ACL2023}

\usepackage{times}
\usepackage{latexsym}

\usepackage[T1]{fontenc}

\usepackage[utf8]{inputenc}

\usepackage{microtype}

\usepackage{inconsolata}
\usepackage{microtype}
\usepackage{graphicx}
\usepackage{multirow}
\usepackage{subfigure}
\usepackage{amsmath, amssymb, mathtools}
\usepackage[export]{adjustbox}
\usepackage{hyperref}
\title{Seen to Unseen: Exploring Compositional Generalization of Multi-Attribute Controllable Dialogue Generation}

\author{Weihao Zeng$^{1*}$, Lulu Zhao$^{1*}$, Keqing He$^{2}$, Ruotong Geng$^{1}$ \\ {\bf Jingang Wang$^{2}$,} {\bf Wei Wu$^{2}$,}  {\bf Weiran Xu$^{1}$}\thanks{\ \ The first two authors contribute equally. Weiran Xu is the corresponding author.}\\
  $^1$Beijing University of Posts and Telecommunications, Beijing, China\\
$^{2}$Meituan, Beijing, China\\
  \texttt{\{zengwh,zhaoll,ruotonggeng,xuweiran\}@bupt.edu.cn}\\
  \texttt{\{hekeqing,wangjingang,wuwei\}@meituan.com}
  }

\begin{document}
\maketitle
\begin{abstract}
Existing controllable dialogue generation work focuses on the single-attribute control and lacks generalization capability to out-of-distribution multiple attribute combinations. In this paper, we explore the compositional generalization for multi-attribute controllable dialogue generation where a model can learn from seen attribute values and generalize to unseen combinations. We propose a prompt-based disentangled controllable dialogue generation model, DCG. It learns attribute concept composition by generating attribute-oriented prompt vectors and uses a disentanglement loss to disentangle different attributes for better generalization. Besides, we design a unified reference-free evaluation framework for multiple attributes with different levels of granularities. Experiment results on two benchmarks prove the effectiveness of our method and the evaluation metric. 
\end{abstract}

\section{Introduction}

Recently, large pre-trained language models (PLMs) like DialoGPT \cite{zhang-etal-2020-dialogpt}, BlenderBot \cite{roller2020recipes} and Meena \cite{adiwardana2020towards} can produce fluent and relevant responses for dialogue contexts. However, the generated responses are often uninformative and factual inconsistent. Hence, controllable dialogue generation (CDG) is proposed to guide dialogue generation towards the desired attributes such as emotions \cite{zhou2018emotional}, acts \cite{li-etal-2017-dailydialog}, and personas \cite{zhang-etal-2018-personalizing}. Previous work focused on directly fine-tuning the large-scale PLMs \cite{keskar2019ctrl} or using an extra attribute discriminator \cite{krause-etal-2021-gedi-generative,dathathri2019plug} to guide generation. The former is expensive and requires extensive annotated attribute labels. The decoding of the latter is computationally intensive, reducing the response fluency and generation speed.

\begin{figure}[t]
    \centering
    \begin{adjustbox}{minipage=\linewidth,scale=1.0}
    \subfigure[E-ACC]{
        \includegraphics[width=0.49\textwidth]{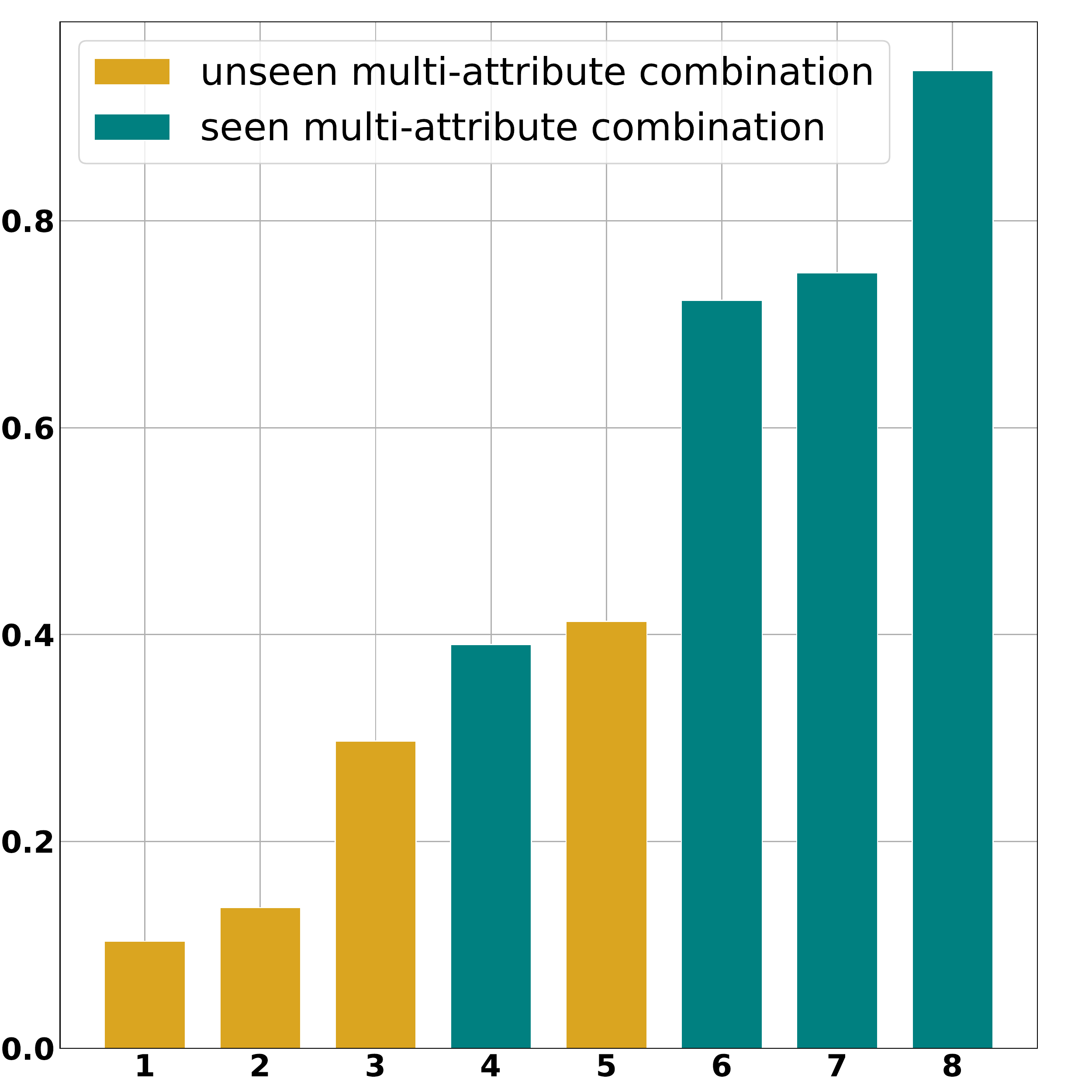}
    }
    \hspace{-0.5cm}
    \subfigure[A-ACC]{
        \includegraphics[width=0.49\textwidth]{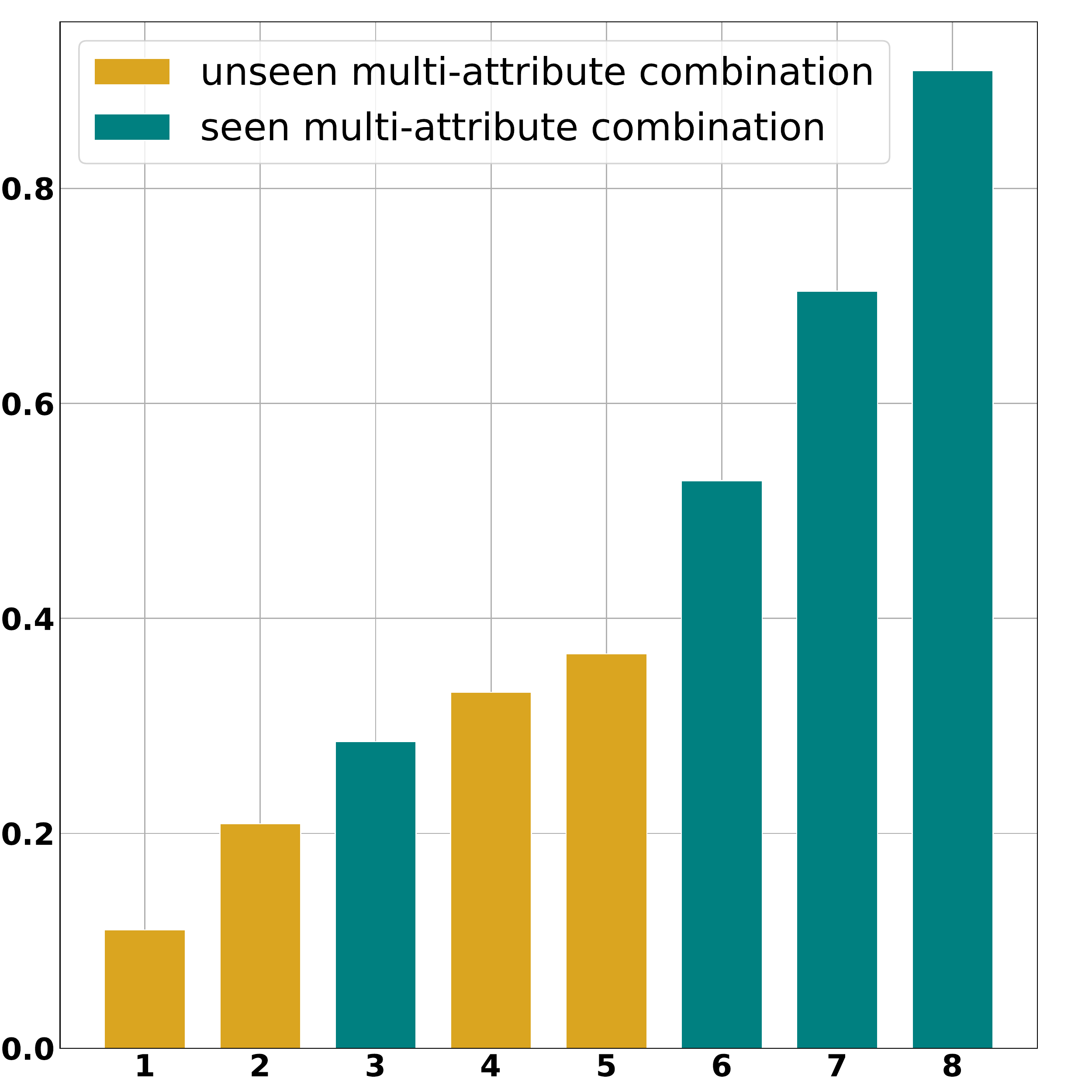}
    }
    \vspace{-0.5cm}
    \caption{The difference of controllability scores on seen and unseen multi-attribute combinations of CTRL \cite{keskar2019ctrl}. E-ACC and A-ACC denote emotion accuracy and act accuracy.}
    \label{fig:seen_unseen_diff}
    \vspace{-0.7cm}
    \end{adjustbox}
    \vspace{-0.3cm}
\end{figure}

Although these methods have made some progress in CDG, most of them focus on single-attribute generation where there is only one attribute label like \emph{happiness} in emotion and pay less attention to the multi-attribute generation, which is a more practical setting. Therefore, we are committed to filling this gap in CDG. Noted that different from single-attribute, the control signal of the multi-attribute generation is a combination of multiple values from different attributes, which faces the challenge of lacking sufficient annotated attribute-specific data. We also find state-of-the-art methods for multi-attribute controllable text generation \cite{yang2022tailor,qian-etal-2022-controllable}, which combine controllers learned from single-attribute, only suitable for discrete attributes with specific labels \cite{li-etal-2017-dailydialog} but not for continuous attributes \cite{zhang-etal-2018-personalizing}. More importantly, we further show directly applying all existing models achieves superior attribute accuracy on seen attribute combinations but drops significantly on unseen combinations, as shown in Figure \ref{fig:seen_unseen_diff}. It proves that previous work lacks compositional generalization capability from seen attribute values to unseen combinations. Besides, the evaluation of controllability in CDG is severely limited by attribute types and annotated attribute data \cite{du-ji-2021-sidecontrol-controlled}, which is not applicable to all cases. Therefore, it is valuable to explore a unified and efficient evaluation metric.

\begin{figure}[t]
\centering
\resizebox{0.47\textwidth}{!}{
\includegraphics[scale=0.5]{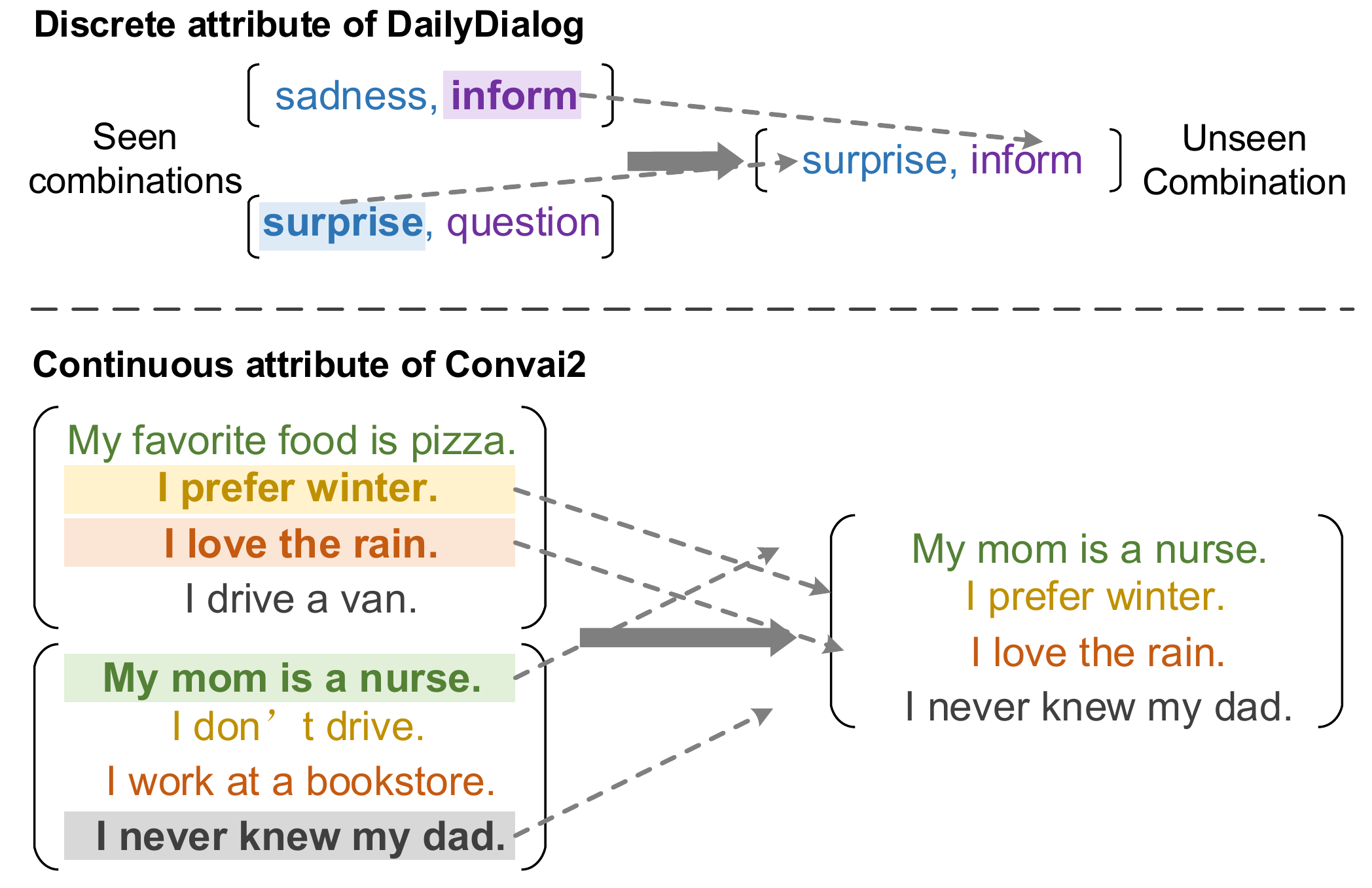}}
\vspace{-0.1cm}
\caption{Examples of the compositional generalization for coarse-grained discrete attributes and fine-grained continuous attributes.}
\label{fig:intro_case}
\vspace{-0.3cm}
\end{figure}

In this paper, we try to explore the compositional generalization for multi-attribute controllable dialogue generation where a model could learn from seen attribute values and generalize to unseen combinations. Figure \ref{fig:intro_case} shows two granularities of multi-attribute compositional generalization, where the token-level attribute labels are regarded as coarse-grained discrete attributes and the sentence-level attribute descriptions are regarded as fine-grained continuous attributes. Specifically, we propose a \textbf{D}isentangled \textbf{C}ontrollable \textbf{G}eneration model (\textbf{DCG}), for compositional generalization in multi-attribute controllable dialogue generation. Inspired by prompt learning \cite{lester-etal-2021-power}, we adopt the attribute values in a combination as attribute-oriented prompts to elicit knowledge from PLMs where the prompts for all instances learn a shared transformation layer, instead of learning an independent prompt representation for each attribute value \cite{clive2022control,qian-etal-2022-controllable,yang2022tailor}. Our method helps transfer attribute concepts from seen values to unseen combinations by learning different prompt embeddings and is easily applied to attribute combination with a huge number of discrete or continuous attribute values. To further disentangle different attribute values, we construct a set of pseudo combinations and design a novel objective of controllable attribute combinations for prompt-tuning, which separates desired attribute combination from others.

Furthermore, to unify the evaluation of different granularity attributes, we design a novel and general reference-free evaluation framework, i.e. \textbf{M}ultiple \textbf{A}ttribute \textbf{E}valuation (\textbf{MAE}), to measure the consistency between desired seen/unseen attribute combinations and generated responses. Specifically, the evaluation of each attribute is converted to a text-to-text generation task based on T5 \cite{raffel2020exploring} with handcrafted templates, and the generated probability of "yes" is regarded as the controllability score. To mitigate the potential bias of different handcrafted modalities \cite{zhao-etal-2019-moverscore,ke-etal-2022-ctrleval}, we add a trainable continuous prompt to improve stability and robustness. Through human evaluation, we show that our proposed evaluation metric can handle both coarse-grained discrete attributes and fine-grained continuous attributes well.

Our contributions are as follows: (1) To the best of our knowledge, we are the first to explore the compositional generalization for multi-attribute controllable dialogue generation and find existing models lack generalization capability to out-of-distribution multi-attribute combinations. (2) We propose a disentangled controllable generation, DCG, which learns attribute concepts from seen values to unseen combinations via a shared mapping of attribute-oriented prompts and uses a disentanglement loss to disentangle different attribute combinations. (3) We introduce a unified reference-free evaluation framework, MAE, for different granularities of attributes. Two benchmarks are established and sufficient experiment results prove the effectiveness of our method and evaluation metric.

\section{Related Work}

\textbf{Controllable Dialogue Generation}  Currently, there have existed many studies on CDG \cite{zhou2018emotional,li-etal-2017-dailydialog,zhang-etal-2018-personalizing}. CTRL \cite{keskar2019ctrl} used 55 kinds of attribute control codes to finetune an LM which is expensive and requires extensive annotated attribute labels. \citet{krause-etal-2021-gedi-generative,dathathri2019plug,yang-klein-2021-fudge,lin2021plug} addressed these limitations by employing an attribute discriminator to update the hidden activations or re-weight the next token distributions, resulting in a slow inference speed. Despite the progress, these models all focus on the single-attribute CDG where the attribute only contains coarse-grained discrete values, such as \emph{happiness} in emotion-controlled generation. It is also vital to explore multi-attribute CDG with multi-granularity attributes. Recently, some works \cite{yang2022tailor,qian-etal-2022-controllable} extend to multi-attribute controllable text generation by simply concatenating the prefixes trained for single attribute. However, they are only suitable for discrete attributes but not for fine-grained continuous attributes like personas \cite{zhang-etal-2018-personalizing}. Besides, we find all these methods have a large performance drop from seen attribute values to unseen combinations. Therefore, in this paper, we are the first to explore the compositional generalization for multi-attribute CDG where a model could learn from seen attributes and generalize to out-of-distribution (OOD) combinations.

\noindent{\textbf{Compositional Generalization in NLP}} Compositional generalization has gradually attracted the interest of NLP researchers. The main application is in semantic parsing, involving grammar-based approaches \cite{herzig-berant-2021-span}, data augmentation strategies \cite{oren-etal-2020-improving}, disentangled representations \cite{zheng-lapata-2022-disentangled}, etc. Recently, a large-scale benchmark, STYLEPTB, is constructed to advance the development of compositional style transfer \cite{lyu2021styleptb}, and a template-based input representation is also performed on the data-to-text task \cite{mehta-etal-2022-improving}. Overall, the application of compositional generalization in NLP tasks is not widespread and there is no related work on CDG at all. 

\noindent{\textbf{Prompt Learning}} Prompt-based methods have achieved significant success in many NLP fields \cite{lester-etal-2021-power,schick-schutze-2021-just}. \citet{li-liang-2021-prefix} proposed the task-specific continuous prompts to finetune a NLG model. For controllable generation, \citet{clive2022control,qian-etal-2022-controllable,yang2022tailor} applied the prompt learning to represent each attribute value as an independent prefix. However, those methods are impractical for fine-grained attributes with a large value set. In contrast, we use the control codes to generate attribute-oriented prompts to guide the generation via a shared MLP layer.

\section{Problem Formulation}

Given a predefined set of attributes $\mathcal{X}=\{A, B, C,...\}$, each attribute contains various values $A=\{a_1,...,a_k\}$ and $k$ is the number of values of attribute $A$. Multi-attribute controlled dialogue response generation aims to generate responses $r$ that satisfy multiple desirable attributes $c=(a_1, b_2,...)$ conditioned on the dialogue history $d$, where $a_1$ and $b_2$ are one value of the attribute $A$ and $B$, and $c\in C_v$ is a combination of attribute values. It can be symbolized as $p(r|d,a_1, b_2,...)$, $(a_1\in A, b_2\in B,...)$.

In this paper, we further focus on the multi-attribute compositional generalization, where the combinations of multiple attribute values for the training set and the test set are disjoint, i.e., $C_{v,train}\cap C_{v,test}=\varnothing$.

\begin{figure}[t]
\centering
\resizebox{0.47\textwidth}{!}{
\includegraphics[scale=0.5]{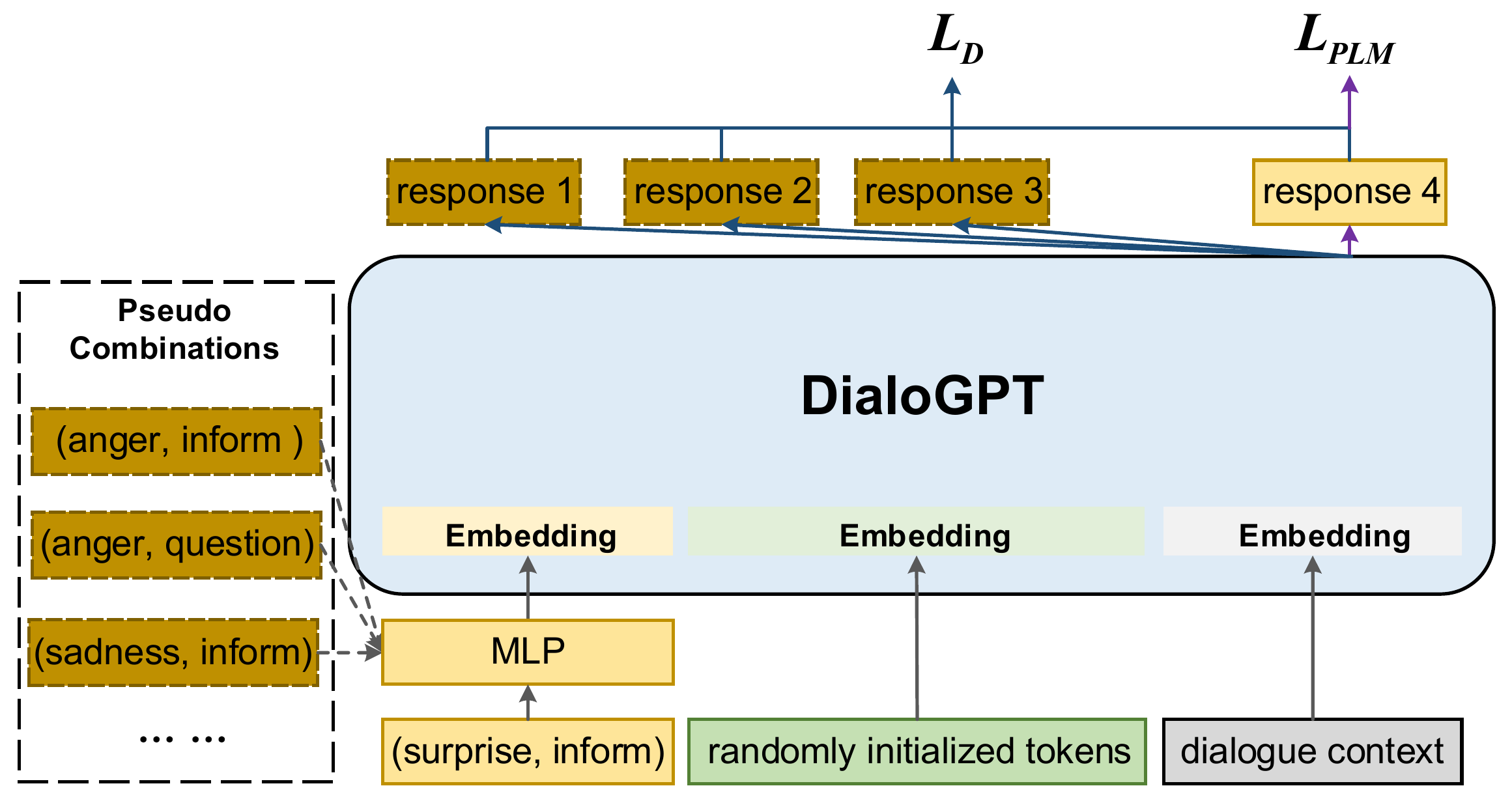}}
\vspace{-0.3cm}
\caption{Overall architecture of our DCG model.}
\label{fig:model}
\vspace{-0.3cm}
\end{figure} 

\section{Methodology}
As shown in Figure \ref{fig:model}, our model is on the basis of the framework of DialoGPT \cite{zhang-etal-2020-dialogpt} with the compositional prompt module.
\subsection{Compositional Prompt}
\subsubsection{Prompt Design}
To better use the control signals, we design two types of prompts to elicit the attribute-related information from the PLM:

\noindent\textbf{Attribute-oriented Prompt} We use the combination of controlled attribute values corresponding to each instance as prompts to guide the model to focus on the controlled information in the dialogue. Here, the controlled attribute values are discrete attribute labels in DailyDialog or continuous attribute descriptions in ConvAI2. The multiple attribute values $a_{i, \cdot}$ in the corresponding combination $c$ are simply concatenated as an attribute-oriented prompt sequence, i.e., $p_{att}=[a_1, b_2,...]$. We encode the prompt tokens using the word embedding layer of a pre-trained DialogGPT and then employ a shared $\operatorname{MLP}_{\theta_1}$ to generate the embeddings $E_{att}$ of the attribute-oriented prompts. Note that we don't require independent parameters for each attribute value like \citet{clive2022control,qian-etal-2022-controllable,yang2022tailor}, but only a shared transformation MLP layer.

\noindent\textbf{Task-oriented Prompt} Although attribute-oriented prompts capture the instance-specific control signals, the dialogue response generation task also is guided by the instance-independent global features. Following \citet{lester-etal-2021-power}, we adopt a series of randomly initialized tokens as the task-oriented prompt, i.e., $p_{task} = [p_1,...,p_m]$, where $m$ is the length of the task-oriented prompt sequence. We look up this prompt sequence in the randomly initialized embedding table $\operatorname{M}_{\theta_2}$ and get the prompt embeddings $E_{task}$.

Finally, we concatenate the two prompt embeddings as the whole prompt embeddings, i.e., $E_{p} = [E_{att};E_{task}]$. 

\begin{table*}[t]
\begin{center}
\resizebox{1.0\textwidth}{!}{
\begin{tabular}{l|c|c|c|c|c|c|c|c|c|c}
\hline 
\bf \multirow{2}{*}{Split} & \multicolumn{5}{c|}{\bf DailyDialog-CG} & \multicolumn{5}{c}{\bf ConvAI2-CG} \\
\cline{2-11}
& \textbf{Size} & \textbf{Turn.num} & \textbf{Att\_com.num} & \textbf{Dial.len} & \textbf{Res.len} & \textbf{Size} & \textbf{Turn.num} & \textbf{Att\_com.num} & \textbf{Dial.len} & \textbf{Res.len} \\
\hline
Train & 12,504 & 6.8 & 18 & 77.6 & 12.9 & 18,000 & 5.0 & 11,566 & 46.5 & 11.7 \\
Validation & 1,390 & 6.5 & 18 & 75.0 & 13.0 & 2,000 & 5.0 & 1,883 & 46.8 & 11.6\\
Test & 1970 & 6.0 & 6 & 69.6 & 13.9 & 2,000 & 5.0 & 873 & 46.1 & 11.6 \\
\hline
\end{tabular}}
\end{center}

\caption{Statistics of DailyDialog-CG and ConvAI2-CG ("CG" means compositional generalization). "Size" and "Att\_com.num"denote the numbers of examples and attribute combinations. "Turn.num" are the average number turns per example. "Dial.len" and "Res.len" are the average lengths of dialogue history and response.}

\label{datasets}
\end{table*}

\subsubsection{Disentanglement Learning} 
Given an instance $(d, c)$, $d$ is the dialogue history and $c$ is the combination of controllable attribute values. To force the model to distinguish different combinations of multiple attribute values, we design some pseudo combinations to enhance the diversity of the prompts, which improves the generalization ability of our model. A disentanglement loss $\mathcal{L}_{D}$ is further introduced to disentangle the combination representations and train multiple compositional prompts simultaneously:
\begin{equation}
\begin{aligned}
\mathcal{L}_D = -log\frac{P(r|d, c)}{P(r|d, c)+\sum_{c^{'}\in C_{pse}}P(r|d, c^{'})} \\
\end{aligned}
\end{equation}
where $C_{pse}$ is the set of pseudo combinations and at least one value in the combination $c^{'}$ is different from the corresponding value in the golden combination.\footnote{We find constructing pseudo combinations with at least one different attribute value is slightly better than with all different attributes in the experiments.} Here, we maximize the generated likelihood of the desirable positive combination $P(r|d, c)$ against the generated likelihood of pseudo combinations $P(r|d, c^{'})$ to generate more controllable responses relevant to given attributes.

\subsection{Training Strategy}

We use DialoGPT \cite{zhang-etal-2020-dialogpt} as the backbone of our model. Given the dialogue history $d$, the embedding $E_d$ is obtained by DialoGPT. Then, the embeddings of the prompt sequence $E_p$ are prepended to the $E_d$ as a whole input embedding matrix. Overall, the PLM loss is calculated as:
\setlength{\abovedisplayskip}{0.1cm}
\setlength{\belowdisplayskip}{0.2cm}
\begin{equation}
\begin{aligned}
\mathcal{L}_{PLM}=-\sum^{T}_{t=1}{\log}\, p_{\theta_1,\theta_2,\varphi}(y_t|y_{<t},d,p_{att},p_{task})\\
\end{aligned}
\end{equation}
where $T$ is the length of generated sequence, i.e., the dialogue history and response. $\varphi$ is the parameter of the PLM and is fixed. The parameters of two prompts, $\theta_1$ and $\theta_2$, are the only updated parameters. Therefore, the training loss $\mathcal{L}$ is the weighted sum of the disentanglement loss and the PLM loss:
\setlength{\abovedisplayskip}{0.1cm}
\setlength{\belowdisplayskip}{0.2cm}
\begin{equation}
\begin{aligned}
\mathcal{L}=\alpha\mathcal{L}_D+(1-\alpha)\mathcal{L}_{PLM}\\
\end{aligned}
\end{equation}

When the training is completed, we save all parameters of the prompt module. During the inference, the data from the test set is mapped to the representations of prompts only via the embedding matrices, where the features of the attributes seen in the training set can be transferred to the unseen combinations.

\begin{figure}[t]
\centering
\includegraphics[width=6.8cm, height=6.5cm]{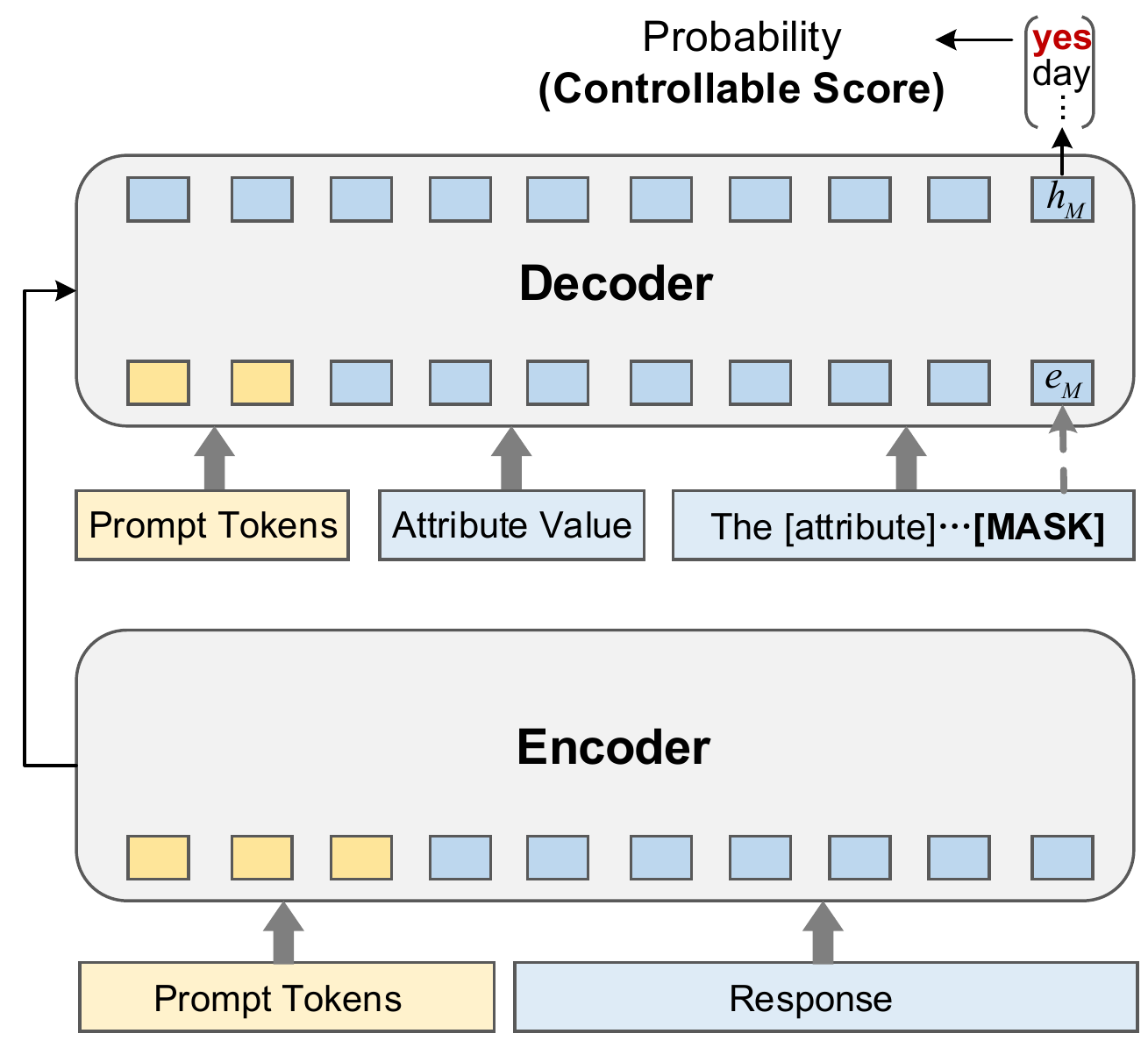}
\vspace{-0.3cm}
\caption{Overview of our evaluation model, MAE.}
\label{fig:model2}
\vspace{-0.7cm}
\end{figure}

\section{Method of MAE}

\label{sec:metric}
To fill the gap in metrics for multi-attribute controllable dialogue generation, we propose a unified and efficient evaluation framework without additional large-scale labeled data, as shown in Figure \ref{fig:model2}, which converts the evaluation of each attribute to a unified text-to-text generation task, just like \citet{gu-etal-2022-ppt}. T5 \cite{raffel2020exploring} is used as the base model for our work. A template is designed as discrete prompts, i.e., "The emotion/act/persona controls the response [MASK]". To alleviate the potential bias of different handcrafted patterns \cite{ke-etal-2022-ctrleval}, we further add a trainable continuous task-oriented prompt to improve stability and robustness.

Specifically, the continuous prompt sequence is prepended to the response as a prefix, which makes up the input of the encoder. Another continuous prompt sequence, the attribute values, and the template are concatenated and fed to the decoder. We take the probability of generating "yes" corresponding to [MASK] token as the controllability score. In training process, only embeddings of continuous prompts are updated and the parameters of T5 are fixed. Note that our model-based evaluation approach gets rid of the reliance on golden response when tested and can be uniformly applied to various granularities of attributes.

\begin{table*}[]
\centering
\small
\resizebox{0.95\textwidth}{!}{
\begin{tabular}{l|cccc|ccc}
\hline
\multicolumn{1}{c|}{{\color[HTML]{000000} \textbf{}}} & \multicolumn{4}{c|}{{\color[HTML]{000000} \textbf{Controllability}}}                      & \multicolumn{3}{c}{\textbf{Text Quality}}                                                                       \\ \cline{2-8} 
\multicolumn{1}{c|}{\textbf{Method}}                  & \textbf{E-ACC} $\uparrow$  & \textbf{E-MAE} $\uparrow$ & \textbf{A-ACC} $\uparrow$ & \multicolumn{1}{c|}{\textbf{A-MAE} $\uparrow$ } & \multicolumn{1}{l}{\textbf{BLEU-1} $\uparrow$ } & \multicolumn{1}{l}{\textbf{BLEU-2} $\uparrow$ } & \multicolumn{1}{l}{\textbf{METEOR} $\uparrow$ } \\ \hline \hline
DialoGPT-Ori                                          & 50.36          & 60.46          & 27.82          & 31.61                              & 11.53                               & 1.58                                & 9.03                              \\
FUDGE                                        & 60.10          & 64.29          & 27.21          & 29.21                              & 12.24                               & 1.13                                & 8.67                              \\
PPLM                                         & 51.57          & 56.87          & 33.60          & 33.71                              & 11.77                               & 1.34                                & 9.26                              \\
CoCon                                             & 52.79          & 59.99          & 29.44          & 34.51                              & 6.91                                & 0.42                                & 11.50                              \\
Fine-tuning                                           & 62.74          & 66.77          & 35.66          & 37.02                              & 21.64                               & 10.19                               & 19.15                              \\
CTRL                                    & 67.34          & 69.55          & 33.50          & 36.15                              & 24.76                               & 11.42                               & 20.45                              \\ \hline
Prompt-tuning                                       & 57.06          & 62.78          & 30.36          & 32.53                              & 19.71                               & 7.36                                & 15.13                              \\
CatPrompt                                    & 60.91          & 66.50           & 36.75          & 38.43                              & 24.07                               & 11.17                               & 20.72                              \\
\hline
\textbf{DCG (ours)}                                          & \textbf{70.66} & \textbf{72.61} & 38.98 & \textbf{41.63}                     & \textbf{26.33}                      & \textbf{14.16}                      & \textbf{24.57}                     \\
DCG w/o AOP (Prompt-tuning)                                       & 57.06          & 62.78          & 30.36          & 32.53                              & 19.71                               & 7.36                                & 15.13                              \\

DCG w/o TOP                                 & 66.80          & 68.02          & \textbf{41.83}          & 41.50                              & 19.18                               & 6.74                                & 15.63                              \\
DCG w/o DL                                   & 60.41          & 64.57          & 38.07          & 39.45                              & 22.45                               & 9.20                                & 19.55                              \\
\hline
\end{tabular}%
}
\vspace{-0.2cm}
\caption{The performance of compositional generalization in multi-attribute controllable dialogue generation for DailyDialog-CG. "E" and "A" denote  controllable attributes of "Emotion" and "Act". "AOP", "TOP", and "DL" mean attribute-oriented prompt, task-oriented prompt, and disentanglement learning. Results are averaged over three random runs. $\uparrow$ means a higher score is better. ($p < 0.01$ under t-test)}
\label{main_results_dailydialog}
\vspace{-0.2cm}
\end{table*}

\begin{table*}[]
\centering
\resizebox{0.88\textwidth}{!}{
\begin{tabular}{l|ccc|ccc}
\hline
\multicolumn{1}{l|}{}                & \multicolumn{3}{c|}{\textbf{Controllability}}                                                                 & \multicolumn{3}{c}{\textbf{Text Quality}}                                                                       \\ \cline{2-7} 
\multicolumn{1}{c|}{\textbf{Method}} & \multicolumn{1}{l}{\textbf{P-SIM} $\uparrow$} & \multicolumn{1}{l}{\textbf{P-NLI} $\uparrow$} & \multicolumn{1}{l|}{\textbf{P-MAE}$\uparrow$} & \multicolumn{1}{l}{\textbf{BLEU-1}$\uparrow$} & \multicolumn{1}{l}{\textbf{BLEU-2}$\uparrow$} & \multicolumn{1}{l}{\textbf{METEOR}$\uparrow$} \\ \hline \hline
DialoGPT-Ori               & 60.16                             & 72.47                             & 23.12                              & 12.33                               & 1.54                                & 8.95                              \\
PPLM              & 59.90                             & 75.98                             & 25.03                              & 13.20                               & 1.65                                & 9.06                              \\
Fine-tuning                & 65.48                             & 69.50                              & 19.21                              & 16.53                               & 2.40                                & 10.96                              \\
CTRL          & 65.20                             & 77.65                             & 26.12                              & 18.39                               & \textbf{3.12}                       & 12.23                              \\ 
\hline
Prompt-tuning                & 64.84                             & 74.30                              & 24.56                              & 17.59                               & 2.60                                & 11.22                              \\
\hline
\textbf{DCG (ours)}                & \textbf{69.03}                    & \textbf{81.20}                    & \textbf{30.42}                     & \textbf{19.55}                      & 2.68                                & \textbf{12.42}                     \\
DCG w/o AOP (Prompt-tuning)                & 64.84                             & 74.30                              & 24.56                              & 17.59                               & 2.60                                & 11.22                              \\
DCG w/o TOP                 & 67.35                             & 78.50                              & 28.44                              & 12.18                               & 1.05                                & 7.61                              \\
DCG w/o DL                & 68.25                             & 79.00                              & 28.53                              & 18.34                               & 2.39                                & 11.63                              \\
 \hline
\end{tabular}
}
\vspace{-0.2cm}
\caption{The performance of compositional generalization in multi-attribute controllable dialogue generation for ConvAI2-CG. "P" denotes controllable attribute of "Persona". Results are averaged over three random runs. $\uparrow$ means a higher score is better. ($p < 0.01$ under t-test)}
\label{main_results_convai2}
\vspace{-0.3cm}
\end{table*}

\section{Experiments}

\subsection{Datasets}

We construct two datasets based on DailyDialog \cite{li-etal-2017-dailydialog} and ConvAI2 \cite{dinan2020second} for compositional generalization in multi-attribute controllable dialogue response generation.

\noindent\textbf{DailyDialog-CG} DailyDialog is an open-domain dialogue dataset with two controllable attributes: emotion and act. Here, we treat the labels of the two attributes as an attribute combination, e.g., (surprise,  inform). For dialogues, each utterance with two attribute labels is regarded as the response and all preceding texts of this utterance are considered as the corresponding dialogue history. In this way, we get 14,879 examples. We count the attribute combinations labeled in all examples, 18 of which are selected as $C_{v, train}$ and the other 6 are $C_{v, test}$. Then, the examples are divided into the training set and test set according to the combination set. We also extract 10\% samples from the training set as the validation set.

\noindent\textbf{ConvAI2-CG} ConvAI2 is a persona-based dialogue dataset in which the persona profile of each dialogue is consisting of 4 or 5 personalized sentences. We treat each sentence as an attribute value and the sentences in the same position belong to the same attribute. The persona profile is regarded as an attribute combination, e.g., ("My mom is my best friend.", "I've four sisters.", "I believe that mermaids are real.", "I love iced tea."). For each dialogue,  we choose the first 4 utterances as the dialogue history and the 5th utterance as the response. Consistent with the processing method of DailyDialog-CG, we select 11,566 combinations as $C_{v, train}$\footnote{The 1,883 combinations of the validation set are included in the 11,566 combinations of the training set.} and the other 873 combinations as $C_{v, test}$. After that, we obtain the corresponding training set, validation set, and test set.

The statistics about the two datasets are shown in Table \ref{datasets}.

\subsection{Baselines}

We compare our methods with several competitive baselines. The common dialogue generation models are included: (1) DialoGPT-Ori \cite{zhang-etal-2020-dialogpt}; (2) FUDGE \cite{yang-klein-2021-fudge}; (3) PPLM \cite{dathathri2019plug}; (4) Cocon \cite{chan2020cocon}; (5) Fine-tuning; (6) CTRL \cite{keskar2019ctrl}. We also implement some prompt-based methods for comparison: (1) Prompt-tuning \cite{lester-etal-2021-power}; (2) CatPrompt \cite{yang2022tailor}. More details can be seen in Appendix \ref{sec:baseline}\footnote{Our code, models and other related resources are publicly available at \href{https://github.com/Zeng-WH/Seen-to-Unseen}{https://github.com/Zeng-WH/Seen-to-Unseen}.}.

\subsection{Evaluation Metrics}

In this work, we focus on evaluating the attribute controllability and text quality for different controllable generation methods. 

\noindent\textbf{Attribute Controllability} It aims to evaluate whether the method can generate responses constrained by multiple attributes successfully.

1. For the control of coarse-grained discrete attributes in DailyDialog-CG, we use the classification accuracy, i.e., E-ACC and A-ACC, for each attribute computed by an independently trained Roberta classifier \cite{liu2019roberta}, respectively.

2. For the control of fine-grained continuous attributes in ConvAI2-CG, we calculate the cosine similarity between the representations of attribute sentences and the generated response, i.e., P-SIM\cite{du-ji-2021-sidecontrol-controlled}. We also evaluate the model by measuring the consistency of attribute sentences with the generated response via a Roberta-based Natural Language Inference (NLI) model, i.e., P-NLI\cite{madotto2019personalizing}.

3. We propose a unified model-based evaluation metric, i.e., MAE, for various granularities of attributes, the details can be seen in Section \ref{sec:metric}. 

\noindent\textbf{Text Quality} We use the BLEUs \cite{papineni-etal-2002-bleu} and METEOR \cite{banerjee-lavie-2005-meteor} to measure the match scores between generated responses and ground-truth references.

\subsection{Main Results}


\noindent\textbf{Results on DailyDialog-CG} Table \ref{main_results_dailydialog} presents the results of controllable dialogue generation about unseen attribute combinations for DailyDialog-CG. \footnote{Our DCG improves text quality and controllability. The BLEUs seem low because we adopt the same calculation as ParlAI \cite{miller-etal-2017-parlai}, which is lower than results in \cite{li-etal-2017-dailydialog} for different smooth functions. } We conduct experiments based on some strong controllable dialogue generation models and novel prompt-based methods. In general, our DCG outperforms all other baselines in terms of attribute controllability and text quality. Compared to CTRL, our model improves by 1.6\%, 2.7\%, 4.1\% in BLEU-1, BLEU-2, METEOR for text quality, and 3.3\%, 3.1\%, 5.5\%, 5.5\% in E-ACC, E-MAE, A-ACC, A-MAE for attribute controllability. We also find the FUDGE and PPLM, two methods based on the decoding strategy, perform poorly here, especially in text quality, which illustrates the incompatibility of these decoding strategies for combinatorial generalization. Besides, as observed, Catprompt is a relatively well-performing prompt-based baseline, but it is still far worse than our method. This is because it directly concatenates all trained single-attribute prompts as the multi-attribute prompt for test. This inconsistency between training and testing stages decreases the performance. Different from these methods, our method optimizes the language modeling loss only based on discrete prompts for attribute combination and continuous task-oriented prompt, which can focus on the features of multiple attributes at the same time also during the training and achieve a better transfer via a learnable mapping. 

Besides, we also concern whether DCG benefits from attribute-oriented prompt, task-oriented prompt, and disentanglement learning. We find that DCG w/o AOP is the same with Prompt-tuning and it performs poorly in attribute controllability, which shows attribute-oriented prompt plays an important role in guiding the model to focus on the controlled information. After removing the task-oriented prompt, the DCG w/o TOP decreases to 19.18\%, 6.74\%, and 15.63\% on text quality, but still maintains high controllability. It proves task-oriented prompt helps improve text quality. We also conduct experiments to prove that TOP can improve text quality when combined with other methods. (See Appendix \ref{sec: top on text qulaity}). Besides, after removing disentanglement learning, the DCG w/o DL drops significantly, which shows disentanglement learning effectively disentangles attribute combinations and improves the ability of compositional generalization.


\begin{figure*}[t]
    \centering
    \subfigure[E-ACC]{
        \includegraphics[width=0.23\textwidth]{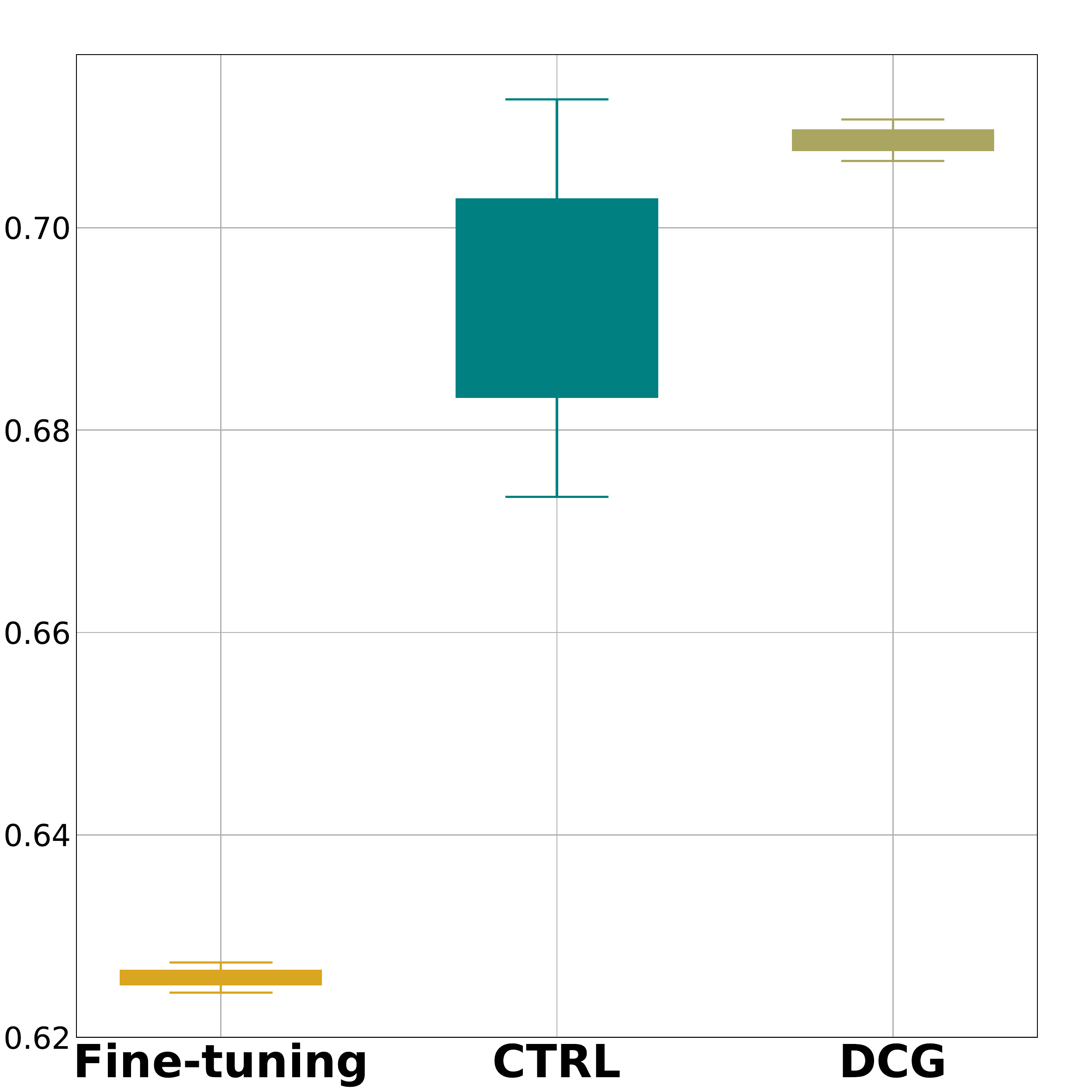}
    }
    \subfigure[A-ACC]{
        \includegraphics[width=0.23\textwidth]{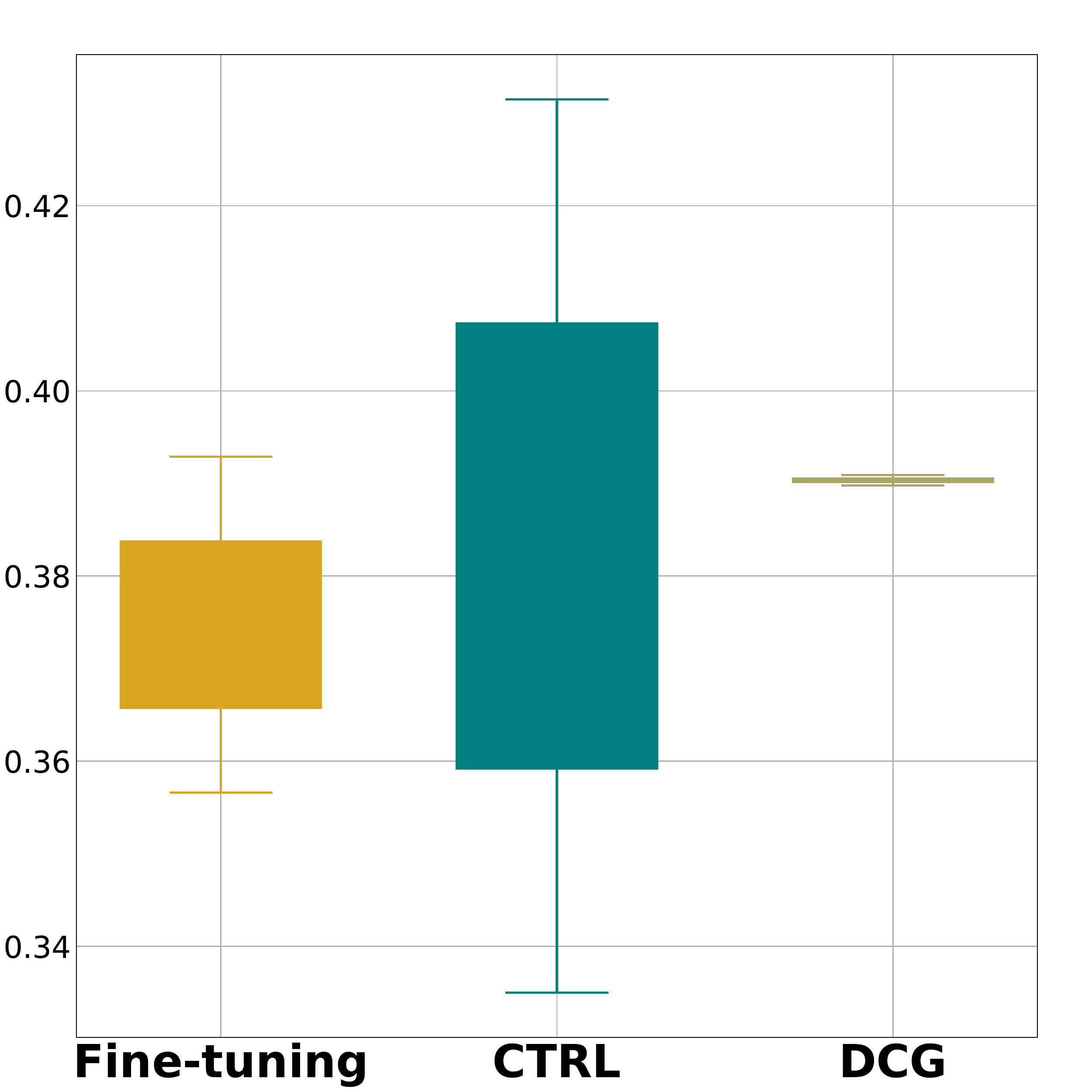}
    }
    \subfigure[BLEU-1]{
        \includegraphics[width=0.23\textwidth]{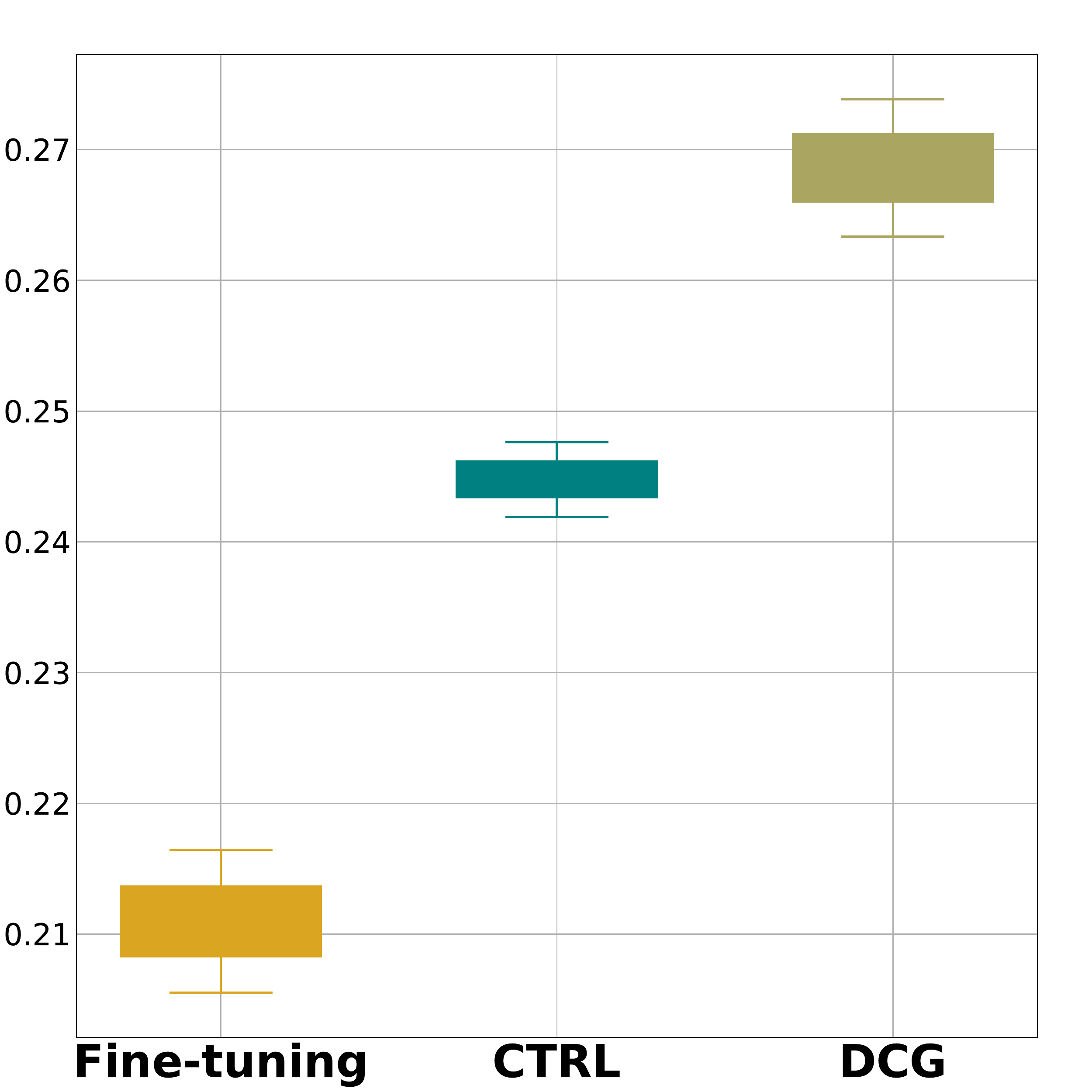}
    }
    \subfigure[BLEU-2]{
        \includegraphics[width=0.23\textwidth]{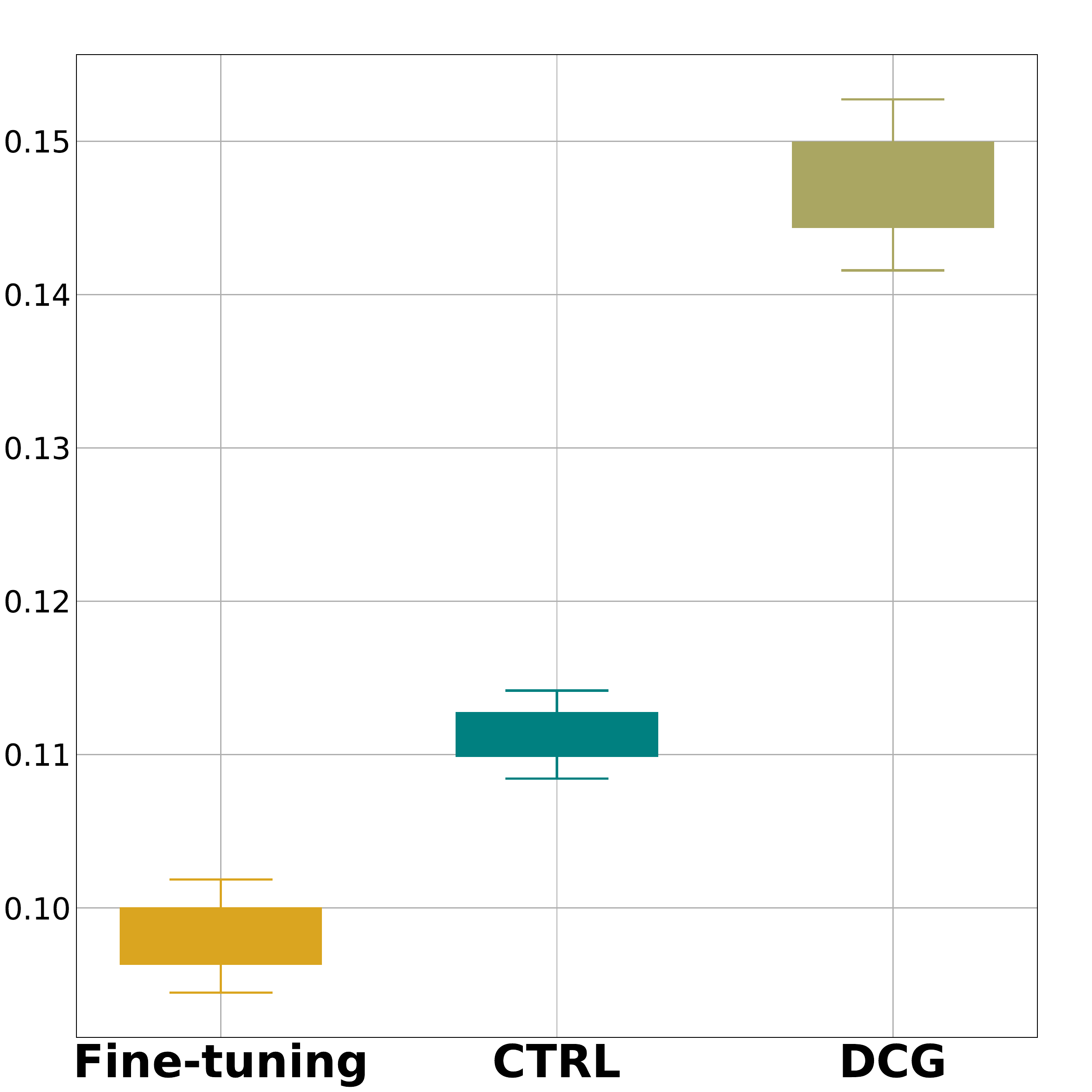}
    }
    \vspace{-0.4cm}
    \caption{Comparision of performance for Fine-tuning, CTRL, and DCG on seen and unseen multi-attribute combinations for DailyDialog-CG in terms of E-ACC, A-ACC, BLEU-1, and BLEU-2.}
    \label{fig:seen_vs_unseen}
    \vspace{-0.1cm}
\end{figure*}

\begin{table*}[htp]
\centering
\resizebox{0.9\textwidth}{!}{
\begin{tabular}{l|ccc|ccc|ccc}
\hline
\multirow{3}{*}{\textbf{Metrics}} & \multicolumn{6}{c|}{\textbf{DailyDialog-CG}} & \multicolumn{3}{c}{\textbf{ConvAI2-CG}} \\
\cline{2-10}
     & \multicolumn{3}{c|}{\textbf{Emotion}}                           & \multicolumn{3}{c|}{\textbf{Act}}                            & \multicolumn{3}{c}{\textbf{Persona}}                            \\ \cline{2-10}
     & \textbf{Pearson}          & \textbf{Spearman}         & \textbf{Kendall}          & \textbf{Pearson}         & \textbf{Spearman}         & \textbf{Kendall}          & \textbf{Pearson}          & \textbf{Spearman}         & \textbf{Kendall}          \\ \hline
ACC           & 0.5242           & 0.4936           & 0.4834           & 0.3852           & 0.4077           & \textbf{0.4027}  & \textbackslash{} & \textbackslash{} & \textbackslash{} \\
P-SIM           & \textbackslash{} & \textbackslash{} & \textbackslash{} & \textbackslash{} & \textbackslash{} & \textbackslash{} & -0.0683          & 0.0065           & 0.0098           \\
P-NLI           & \textbackslash{} & \textbackslash{} & \textbackslash{} & \textbackslash{} & \textbackslash{} & \textbackslash{} & -0.0881          & -0.0741          & -0.0706          \\
\hline
MAE           & 0.6821           & \textbf{0.7500}  & \textbf{0.6242}  & 0.5446           & \textbf{0.4661}  & 0.3936           & \textbf{0.5793}  & 0.5768  & 0.4418  \\ 
MAE w/o Prompt & 0.3665           & 0.4802           & 0.3857           & -0.2832          & -0.2136          & -0.1789          & -0.0529          & 0.2591           & 0.2062           \\
MAE (BART)      & \textbf{0.6829}  & 0.7396           & 0.6102           & \textbf{0.5478}  & 0.4358  & 0.3697           & 0.5550 & \textbf{0.5848}
 & \textbf{0.4517} \\
 \hline
 MAE (T1)                                          & 0.6801           & 0.7661            & 0.6382           & 0.5557           & 0.4661            & 0.3935           & 0.6037           & 0.6235            & 0.4811           \\
MAE (T2)                                          & 0.6758           & 0.7070            & 0.5851           & 0.5357           & 0.4055            & 0.3458           & 0.5724           & 0.5767            & 0.4418           \\
MAE w/o Prompt (T1)                                          & 0.1158           & 0.1053            & 0.0912           & -0.3035           & -0.2684            & -0.2266           & 0.0835           & 0.0984            & 0.0884           \\
MAE w/o Prompt (T2)                                          & 0.0417           & -0.0257            & -0.0210           & -0.2680           & -0.1040            & -0.0835           & -0.0512           & -0.0199           & -0.0295           \\
\hline
\end{tabular}
}
\vspace{-0.1cm}
\caption{Pearson ($r$), Spearman ($\rho$), and Kendall ($\tau$) correlations of attribute controllability evaluation metrics on DailyDialog-CG and ConvAI2-CG. "T1" and "T2" denote the Template 1 and Template 2.}
\label{tab:consistency}
\vspace{-0.3cm}
\end{table*}

\noindent\textbf{Results on ConvAI2-CG} Table \ref{main_results_convai2} presents the results of generalization on unseen attribute combinations for ConvAI2-CG. Due to the diversity of attribute values and attribute combinations, it is very difficult to implement CatPrompt in ConvAI2-CG. Therefore, we remove this baseline. We also remove FUDGE and Cocon for their poor generation quality and slow decoding speed, which is shown in Table \ref{main_results_dailydialog} and Table \ref{tab:decoding_speed}. We can observe that the trend of overall performance is consistent with that of DailyDialog-CG. Compared to CTRL,  our model achieves a great improvement in attribute controllability and text quality, which proves the generality of our methods on the coarse-grained discrete attribute control and fine-grained continuous attribute control. It also shows the effectiveness of our method when more attributes are combined. However, all BLEU scores are low, which is because the ConvAI2-CG has more diverse and complex attribute combinations and leads to the instability of models facing new attribute combinations. Generally, the results show that the compositional generalization for multi-attribute controllable dialogue generation is necessary and meaningful. Noted that we also conduct experiments on the setting with changed number of attributes from training to inference (See in Appendix \ref{sec:Performance on Number of Attribute}).

\section{Qualitative Analysis}

\subsection{Comparison between Seen and Unseen Attribute Values}
\label{seen vs. unseen}
Figure \ref{fig:seen_vs_unseen} displays the comparison of the performance on seen and unseen attribute combinations for DailyDialog-CG. We report the controllability metrics, E-ACC (emotion) and A-ACC (act), and the BLEUs of the Fine-tuning, CTRL, and our DCG. The top of each box denotes the result of seen attribute combinations and the bottom represents unseen attribute combinations. We find all methods achieve significantly superior performance on seen attribute combinations than on unseen combinations. For example, CTRL achieves 71.27\% E-ACC and 43.15\% A-ACC on seen attribute combinations but drops to 67.34\%(-3.93) and 33.50\%(-9.65) on unseen combinations. It strongly proves previous methods suffer from the difficulty of compositional generalization for the multi-attribute controllable dialogue generation. However, we find our proposed DCG can greatly alleviate this gap. The DCG has a smaller drop of 0.41\% and 0.11\% for E-ACC and A-ACC, and it also outperforms CTRL on both controllability and text equality of unseen attribute combinations. The results confirm the effectiveness of our method for transferring seen attributes to unseen combinations. We find CTRL achieves a higher A-ACC on seen combinations but a lower score on unseen combinations than Fine-tuning, which demonstrates directly adding control codes may cause overfitting to seen attribute combinations.

\subsection{Correlation Results on Metrics}

Following \citet{guan-huang-2020-union}, we adopt Pearson ($r$), Spearman ($\rho$), and Kendall ($\tau$) correlation coefficients between our proposed automatic metric, MAE, and human judgments (details can be seen in Appendix \ref{sec:human_evaluation}) to measure the quality of different metrics. Table \ref{tab:consistency} shows the overall results on the controllability of coarse-grained discrete attributes, emotion and act, and the fine-grained continuous attributes, persona description. We can observe that our MAE outperforms classic metrics,  E-ACC, A-ACC, P-SIM, and P-NLI, by a large margin, indicating the effectiveness of our unified metric on different granularities. We also conducted experiments on some variants of MAE. After the removal of continuous prompts, the correlation scores decrease. It is because the task-oriented prompts are the only parameters can be fine-tuned, which is important for MAE. We also implement MAE on another PLM, BART, to demonstrate generality for our model.

\noindent\textbf{Robustness Analysis} To verify the effect of the bias of the handcrafted template, we design another two templates. The Template 1 is  "The response is related to the emotion/act/persona [MASK]" and Template 2  is "The response is about the emotion/act/persona [MASK]". As shown in Table \ref{tab:consistency},  MAE (T1) and MAE (T2) achieve similar correlation results (within 0.50\%) while the results of MAE w/o Prompt (T1) and MAE w/o Prompt (T2) are quite different. It suggests the trainable continuous task-oriented prompt can alleviate the potential bias of different handcrafted templates and further improve the robustness of MAE.

\begin{figure*}[t]
    \centering
    \subfigure[CatPrompt]{
        \includegraphics[width=0.31\textwidth]{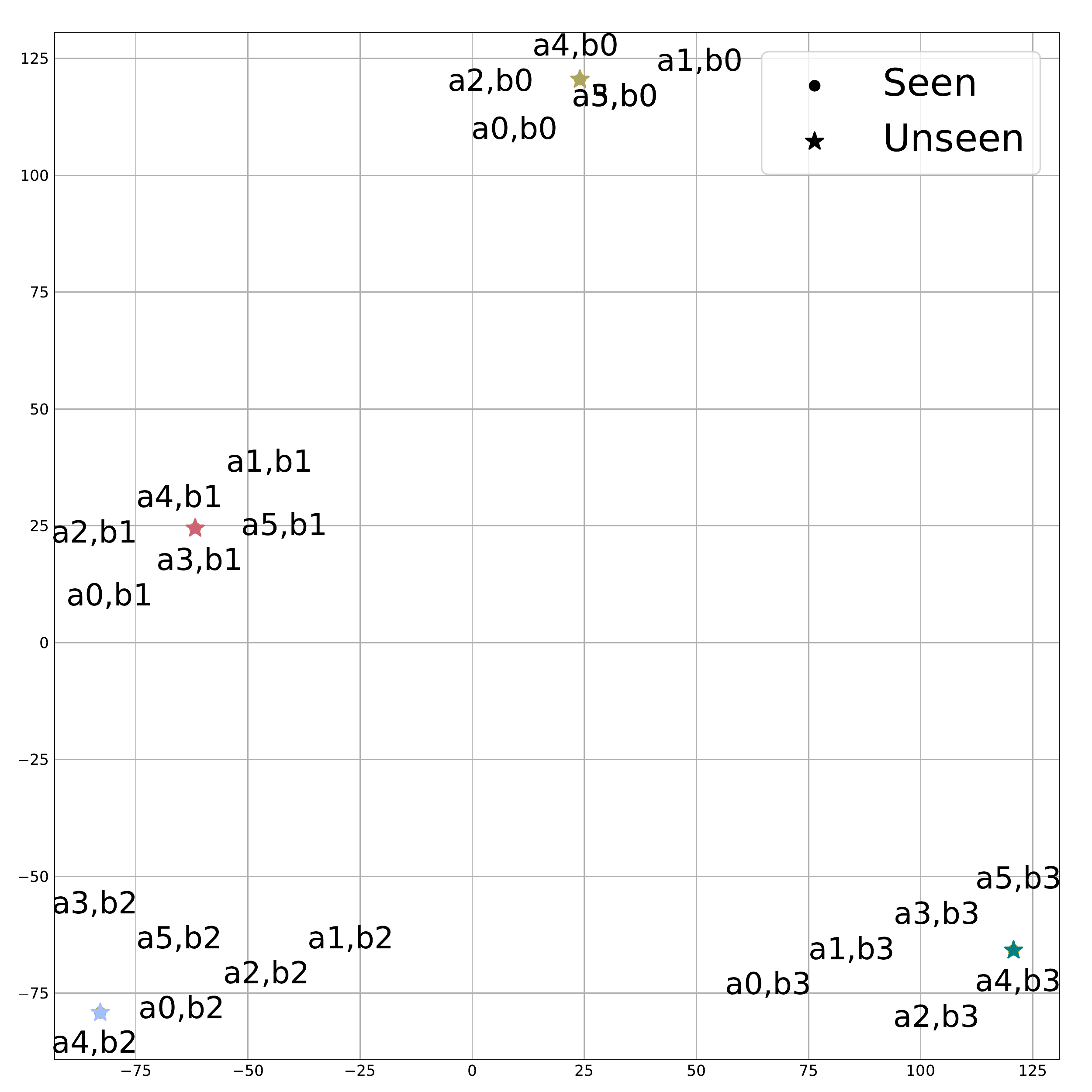}
    }
    \subfigure[CtrlPrompt]{
        \includegraphics[width=0.31\textwidth]{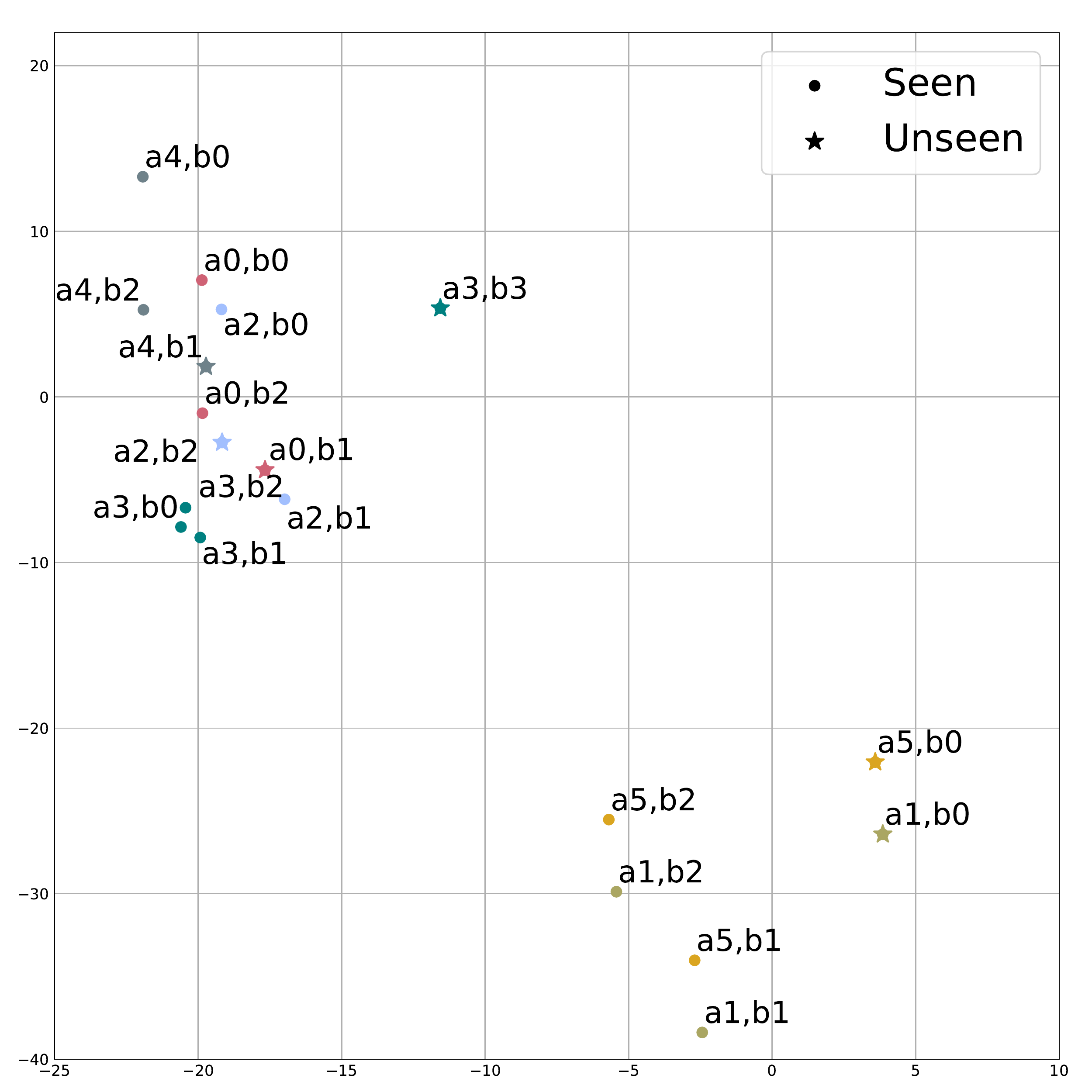}
    }
    \subfigure[DCG]{
        \includegraphics[width=0.31\textwidth]{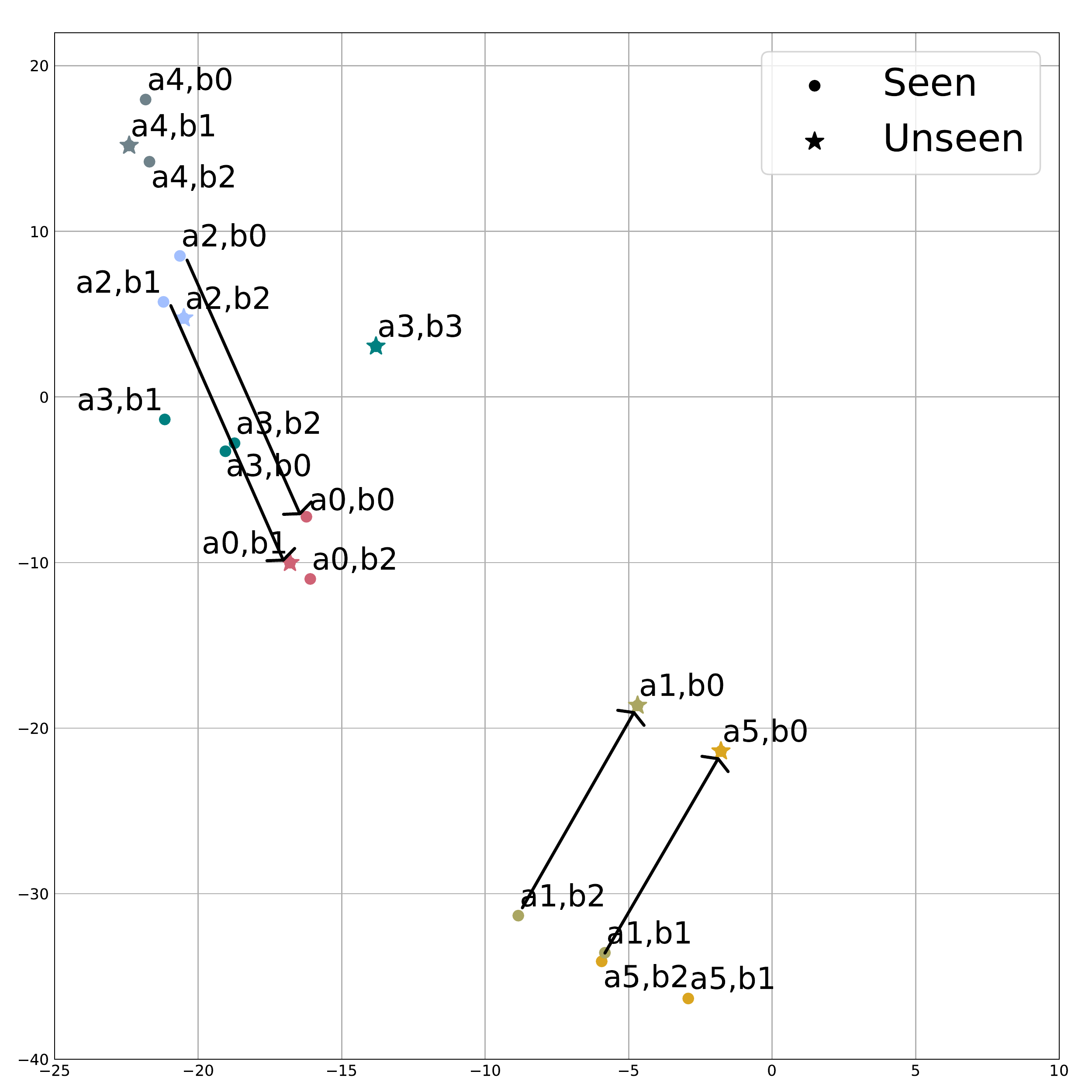}
    }
    \vspace{-0.4cm}
    \caption{Visualization of prompts from different models on DailyDialog-CG. Each dot denotes the prompt embeddings of a multi-attribute combination $(a*, b*)| a* \in A, b* \in B$, where A is the attribute \emph{Emotion} and B is the attribute \emph{Act}. The same color represents that two dots have the same value of \emph{Emotion} and different shaped dots represent seen/unseen combinations. For clarity, we leave out some outliers.}
    \label{fig:pca}
    \vspace{-0.5cm}
\end{figure*}

\begin{figure}[t]
    \centering
    \begin{adjustbox}{minipage=\linewidth,scale=0.98}
    \subfigure[E-ACC]{
        \includegraphics[width=0.47\textwidth]{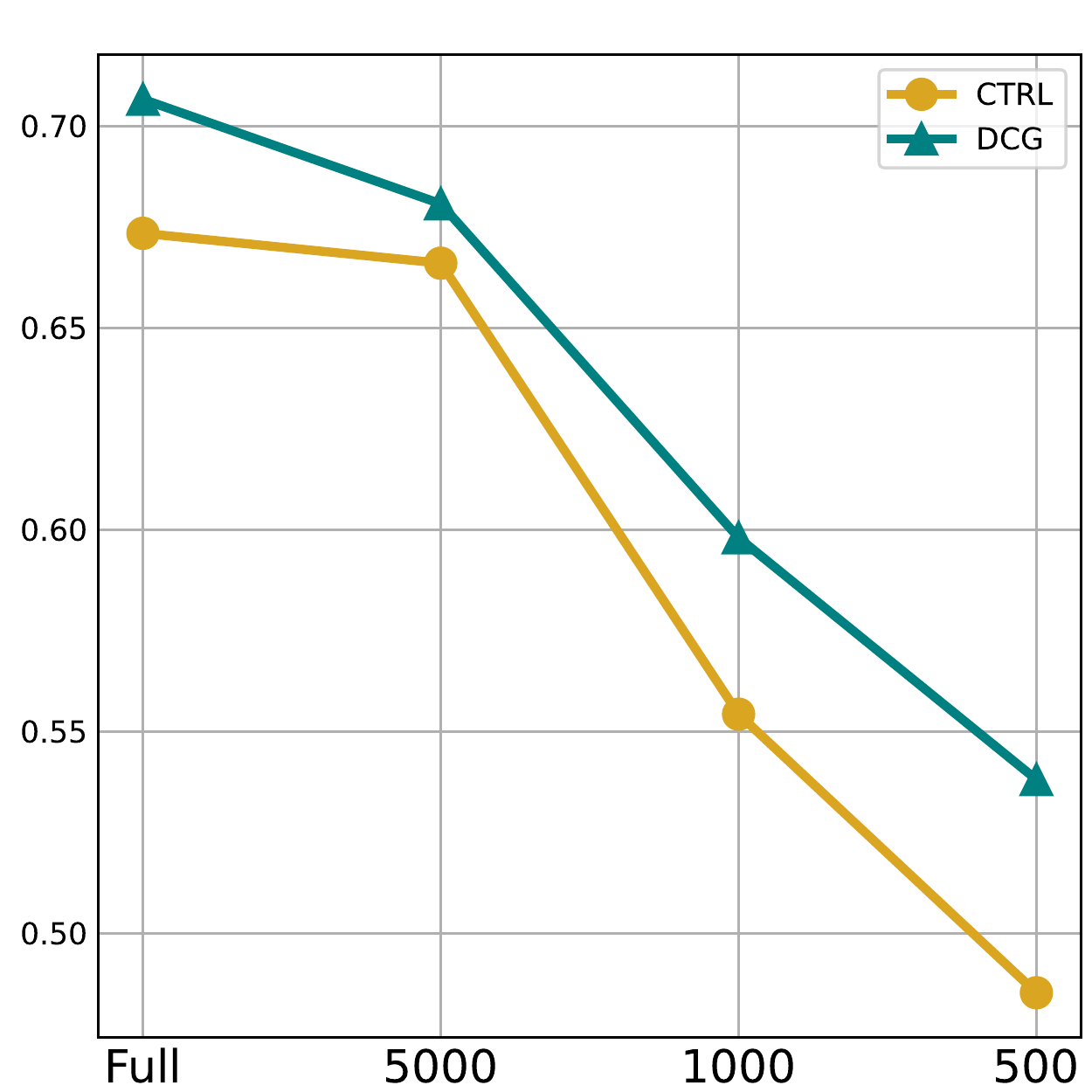}
    }
    \vspace{-0.42cm}
    \subfigure[A-ACC]{
        \includegraphics[width=0.47\textwidth]{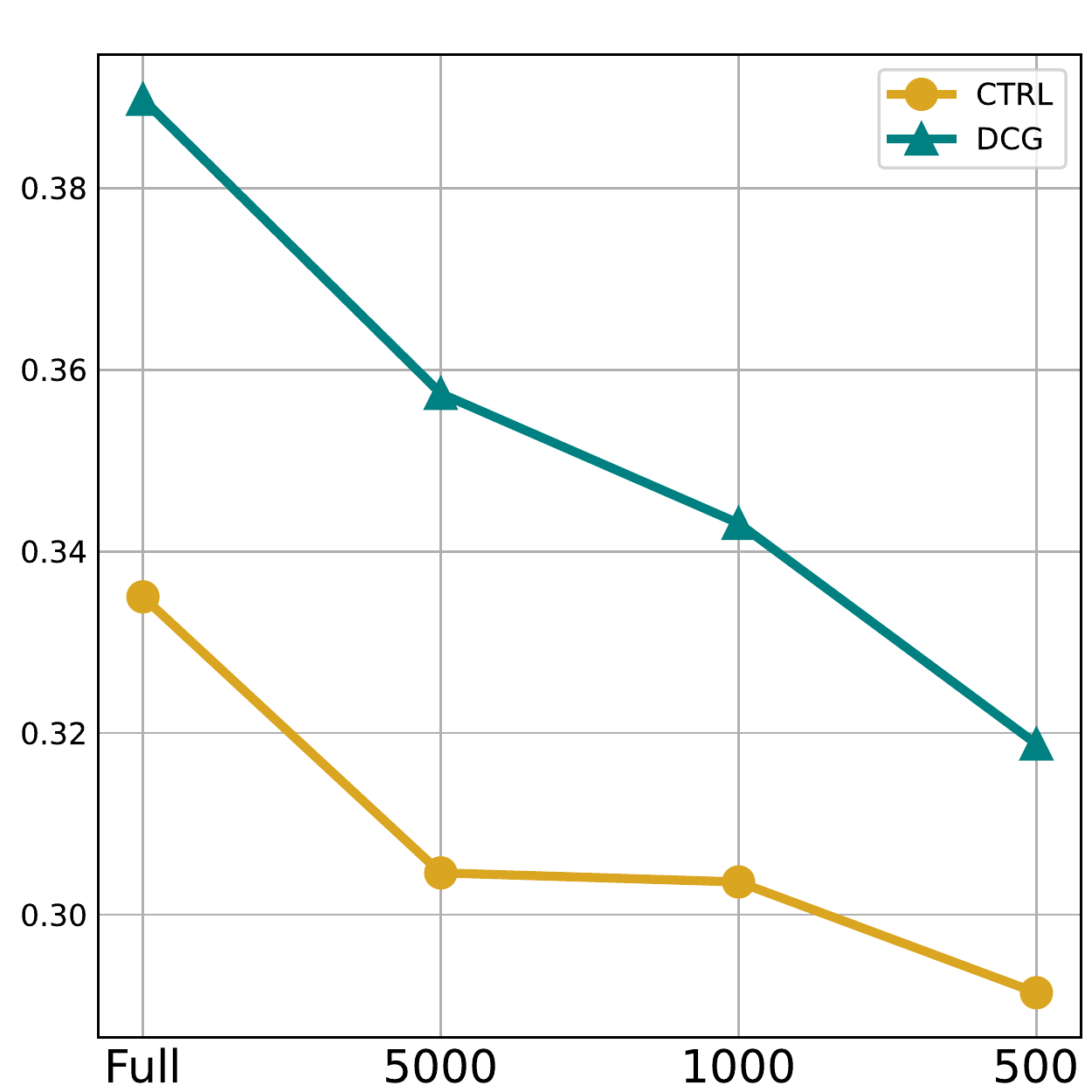}
    }
    \subfigure[BLEU-1]{
        \includegraphics[width=0.47\textwidth]{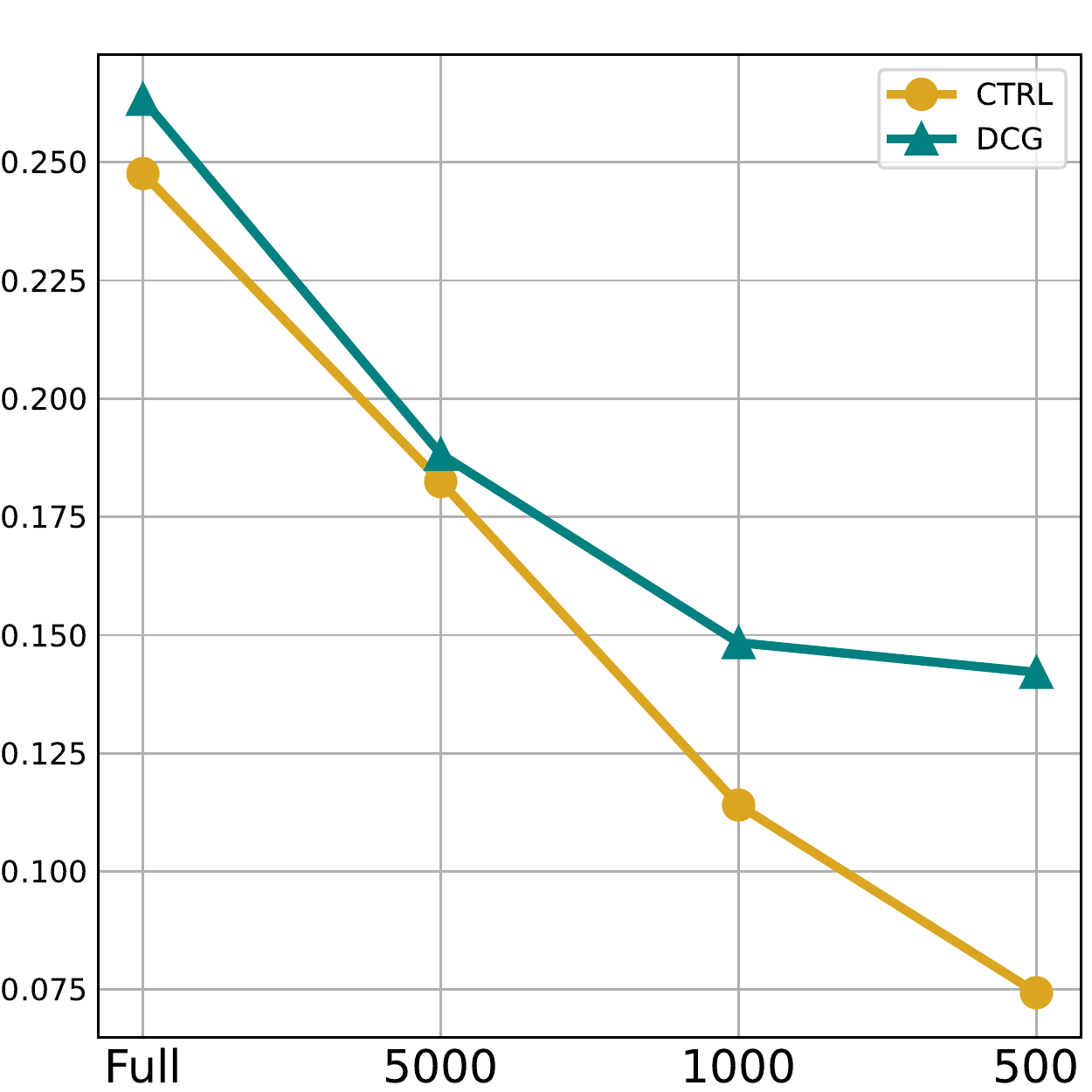}
    }
    \subfigure[BLEU-2]{
        \includegraphics[width=0.47\textwidth]{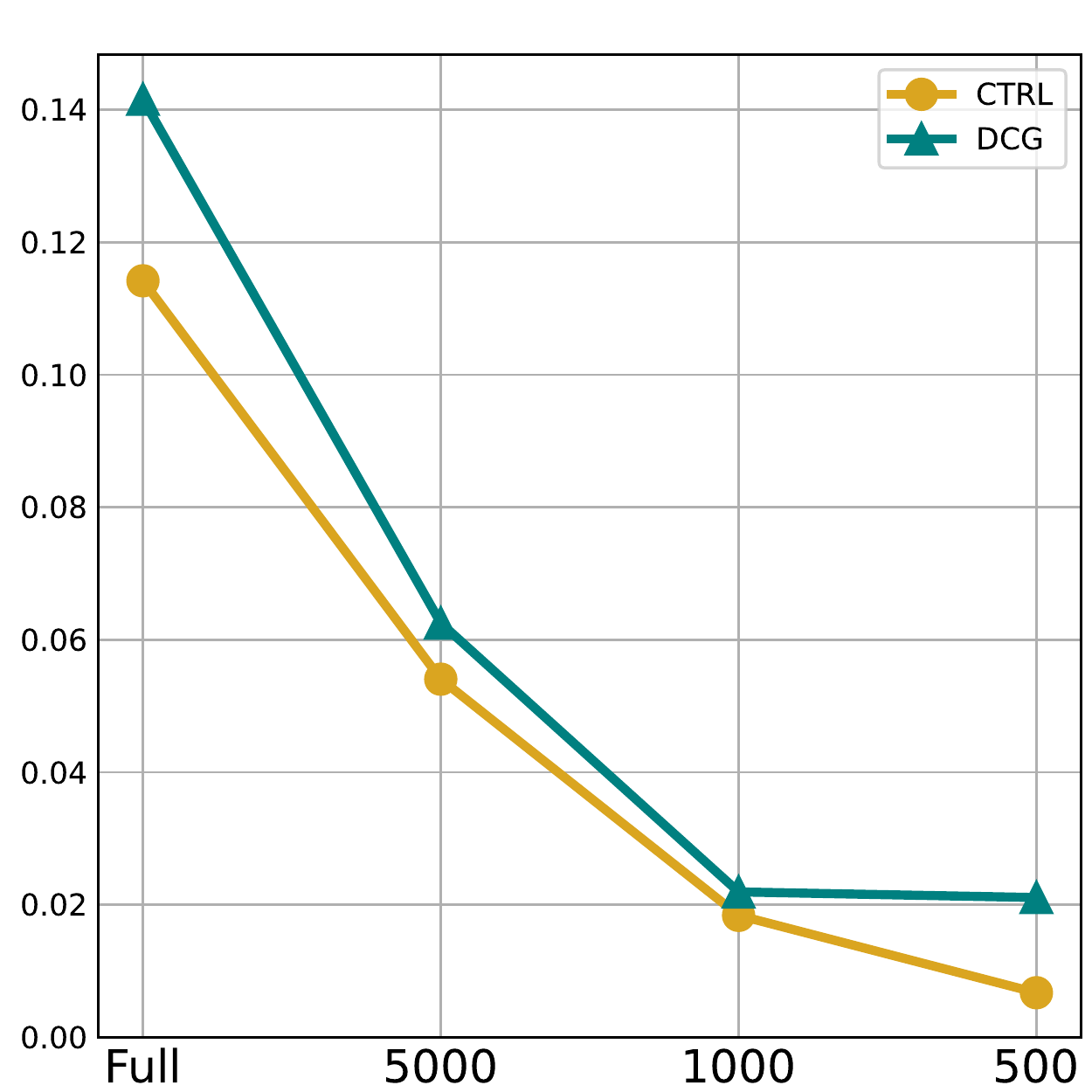}
    }
    \vspace{-0.5cm}
    \caption{Few-shot learning of our DCG and CTRL on DailyDialog-CG.}
    \label{fig:few}
    \end{adjustbox}
    \vspace{-0.3cm}
\end{figure}

\subsection{Prompt Visualization}

To show the effect of prompts for compositional generalization, we display a visualization of the concatenated prompt embeddings of two attributes via PCA \cite{jolliffe2016principal} on DailyDialog-CG in Figure \ref{fig:pca}. For CatPrompt in Figure \ref{fig:pca}(a), all the multi-attribute combinations ($6 (emotion) \times 4 (act) = 24$) almost collapse into four dots where each dot is of the same act attribute value but of different emotion values.  We find directly concatenating two single-attribute prompts makes the model only focus on the latter attribute (act), i.e., position sensitive, so that the CatPrompt cannot distinguish different combinations with the other attribute (emotion). Therefore, it's hard for CatPrompt to learn multi-attribute compositional generalization. In Figure \ref{fig:pca}(b), We find that DCG w/o DL can distinguish different multi-attribute combinations to some extent. However, the combinations of different attribute values are tightly entangled, such as (a0, b2) and (a4, b1). Figure \ref{fig:pca}(c) shows that our DCG has a close distribution with prompts of the same attribute value, i.e., {(a0, b0), (a0, b1), (a0, b2)}, and a sparse distribution with prompts of different attribute values, e.g., (a0, b2) and (a4, b1). It proves our DCG can disentangle attribute combinations and learn relations between different attributes. Furthermore, DCG learns generalization capability from seen attributes to unseen combinations. For example, (a2, b1) -> (a0, b1) (unseen path) is equal to (a2, b0) -> (a0, b0) (seen path). The results confirm that our proposed attribute-oriented prompt outperforms the models that learn an independent prompt for each attribute value. The shared embedding mapping helps learn attribute concepts from seen values to unseen combinations.

\subsection{Few-shot Learning}


To study the effect of few-shot learning, we randomly select a ratio of original training data from DailyDialog-CG to train CTRL or DCG in low-resource settings and evaluate the model performance on the original test set. "Full" denotes the same setting as the main results. 5000, 1000, and 500 denote the number of examples chosen from the original training data respectively. The results are shown in Figure \ref{fig:few}. Note that we keep the original test set fixed for a fair comparison. As the size of training data decreases, the performance of both CTRL and DCG presents a dropping trend and our DCG model is consistently better than CTRL, which confirms our model has a strong capability for multi-attribute controllable dialogue generation.

\section{Case Study}
Figure \ref{fig:case} (See in Appendix) shows two examples from Dailydialog-CG and ConvAI2-CG, respectively. For example one in the DailyDialog-CG, the CTRL generates the word "great", showing that the generated response is emotionally controllable. However, both sentences in the response are declarative sentences, which does not control the act \emph{question}. As observed, the response generated by our DCG contains the word "Wow", which strongly expresses the emotion of \emph{happiness}. Besides, a question sentence is also generated. Example two in ConvAI2-CG needs to control 5 attributes, of which the golden response contains 2 attributes. The CTRL only controls "like to skate", while our DCG controls "like to write poetry and skate", which is highly consistent with the golden response. Compared with previous models, our model addresses many difficult issues in compositional generalization for multi-attribute controllable dialogue generation. With an attribute-oriented prompt and a task-oriented prompt, our method learns attribute concepts from seen attribute values to unseen attribute combinations. Through a disentanglement learning, some artificial-constructed unseen pseudo combinations are injected into the training process, which greatly improves the generalization ability of our model.

\section{Conclusion}

In this paper, we study the compositional generalization for multi-attribute controllable dialogue generation. We propose a prompt-based disentangled controllable dialogue generation model which generates attribute-specific prompt vectors from control codes and uses a disentanglement loss to disentangle different attributes. Further, we develop a unified reference-free evaluation framework, MAE, for multi-attribute generation with different levels of granularities. Experiments and analysis show our method achieves better text quality and controllability scores. Moreover, our proposed MAE has a higher correlation with human judgments for evaluation on CDG.

\section*{Acknowledgements}
We thank all anonymous reviewers for their helpful comments and suggestions. We are also grateful to the track organizers for their valuable work. This work was partially supported by National Key R\&D Program of China No. 2019YFF0303300 and Subject II No. 2019YFF0303302, DOCOMO Beijing Communications Laboratories Co., Ltd, MoE-CMCC "Artifical Intelligence" Project No. MCM20190701. Jingang Wang is funded by Beijing Nova Program(Grant NO. 20220484098)

\section*{Limitations}
Although DCG achieves significant improvements compared with existing baselines, there are still avenues to be explored in future research. (1) DCG in this paper focuses on the compositional generalization for multi-attribute on controllable dialogue generation. We hope to extend the method to other generative tasks, including but not limited to dialogue summarization and story generation. (2) In this paper, we explored the control of coarse-grained discrete attributes and the control of fine-grained ones separately, and we intend to study the combination of these two attributes in future research.

\section*{Ethics Statement}
Controllable dialogue generation(CDG) is an essential task in Natural Language Processing (NLP) and has been widely studied for decades, which aims to guide dialogue generation toward the desired attributes such as emotions, acts, and personas. In the open-domain dialogue scenario, CDG can generate emotional and diverse responses to enhance the user's sense of participation. In the task-oriented dialogue scenario, CDG can generate responses that meet the user's needs according to the user's intent. However, most previous works focus on single-attribute generation where there is only one attribute label like \emph{happiness} in emotion and pay less attention to the multi-attribute generation, which is a more practical setting. Different from single-attribute, the control signal of the multi-attribute generation is a combination of multiple values from different attributes, which faces the challenge of lacking sufficient annotated attribute-specific data. Therefore, we explore the compositional generalization for multi-attribute controllable dialogue generation where a model could learn from seen attribute values and generalize to unseen combinations. We also design a novel and general reference-free evaluation framework to unify the evaluation of different granularity attributes. The experimental results prove the effectiveness of our model and evaluation framework. Besides, there is no huge biased content in the datasets and the models. If the knowledge base is further used, the biased content will be brought into the generated responses, just like biased content posted by content creators on the Web which is promoted by a search engine. To prevent the technology from being abused for disinformation, we look forward to more research effort being paid to fake/biased/offensive content detection and encourage developers to carefully choose the proper dataset and content to build the knowledge base.

\bibliography{anthology,custom}
\bibliographystyle{acl_natbib}

\appendix

\section{Baselines}
\label{sec:baseline}
\noindent\textbf{DialoGPT-Ori}: Proposed by \cite{zhang-etal-2020-dialogpt}, this model is a dialogue generative pre-trained transformer. Here, we use the original DialoGPT for open-domain dialogue generation. DialoGPT is the backbone for all other baselines except CoCon. 

\noindent\textbf{Fine-tuning}: We use dialogue history in datasets to fine-tune the DialoGPT for dialogue generation.

\noindent\textbf{CTRL}: Proposed by \cite{keskar2019ctrl}, this method provides attribute control codes for a language model trained from scratch. We concatenate multi-attribute control codes with dialogue history to fine-tune the DialoGPT. 

\noindent\textbf{CoCon}: Proposed by \cite{chan2020cocon}, this method uses a content input to control an GPT's output text at a fine-grained level.

\noindent\textbf{PPLM}: Proposed by \cite{dathathri2019plug}, this method is a gradient-based baseline that uses a plug-and-play language model(PPLM) to guide the language model. We train a joint classifier of emotion and dialogue act which takes a single response as input and predicts the attribute combination of the emotion and dialogue act on DailyDialog-CG. Noted that the attribute classifiers of PPLM can not directly generalize to unknown attribute combinations, so we use both training data and test data to train the attribute classifiers. We use the bag-of-words attribute model which encodes persona profile to control the DialoGPT on ConvAI2-CG. 

\noindent\textbf{FUDGE}: Proposed by \cite{yang-klein-2021-fudge}, this method is a weighted decoding baseline which uses a future discriminator for generation(FUDGE) to guide the DialoGPT. We train a joint discriminator that takes the dialogue history and the current response as input and predicts the attribute combination of emotion and dialogue act on DailyDialog-CG.

\noindent\textbf{Prompt-tuning}: Proposed by \cite{lester-etal-2021-power}, this method uses continue prompts to fine-tune language models. We apply this method to the DialoGPT for dialogue generation.

\noindent\textbf{CatPrompt}: Inspired by \citet{yang2022tailor,qian-etal-2022-controllable}, we initialize an unique prompt for each single attribute value and concatenate single-attribute prompts as the multi-attribute prompts. We fine-tune multi-attribute prompts for dialogue generation. Note that CatPrompt is only applied to coarse-grained discrete attributes like emotion and act instead of persona. Because persona has a large value set, resulting in numerous parameters (see Table \ref{tab:trainable_parameters}).

\section{Implementation Details}
Our implementation is based on the Hugging Face Transformer models\footnote{https://github.com/huggingface/transformers}. $\operatorname{{DialoGPT}_{Small}}$ is used as a backbone and the input sequence length is truncated to 512 tokens. Following the HuggingFace default setup, we use an AdamW optimizer \cite{loshchilov2017decoupled} and a linear learning rate scheduler with an initial rate of $7.5\cdot10^{-5}$, and the batch size is set to 8. The prompt lengths are set to 50 and 150, the attribute-oriented prompt lengths are set to 6 and 100, the disentanglement loss weight is set to 0.1 and 0.03, and the number of Pseudo Combinations is set to 8 and 6 for DailyDialog-CG and ConvAI2-CG, respectively. Our model is trained on Tesla V100 machines, taking 24 minutes per epoch on DailyDialog-CG and 36 minutes per epoch on ConvAI2-CG. For all experiments, we set the number of training epochs to 30. At the decoding phase, we use a greedy search and max generated tokens of 150.

\begin{table}[t]
\centering
\resizebox{0.3\textwidth}{!}{
\begin{tabular}{l|c}
\hline
\textbf{Method}  & \textbf{Decoding Speed $\uparrow$} \\ \hline
DialoGPT-Ori     & 1.1837x                \\
FUDGE            & 0.0041x                \\
PPLM             & 0.0006x                \\
CoCon            & 0.0044x                \\
Fine-tuning      & 1.1347x                \\
CTRL             & 1.1673x                \\
\hline
Prompt-tuning    & 1.0000x                \\
CatPrompt        & 1.0408x                \\
\hline
DCG (ours) & 1.0490x                \\
DCG w/o DL & 1.0122x                \\
\hline
\end{tabular}
}
\caption{The decoding speed of different models, which takes the decoding speed of the model relative to the Prompt as a metric.}
\label{tab:decoding_speed}
\vspace{-0.cm}
\end{table}

\begin{table*}[]
\centering
\resizebox{0.9\textwidth}{!}{
\begin{tabular}{l|cc|cc}
\hline
\multicolumn{1}{c|}{\multirow{2}{*}{\textbf{Model}}} & \multicolumn{2}{c|}{\textbf{DailyDialog-CG}}                                     & \multicolumn{2}{c}{\textbf{ConvAI2-CG}}                                          \\ \cline{2-5} 
\multicolumn{1}{c|}{}                                 & \multicolumn{1}{c|}{\textbf{Traninable Parameters}} & \textbf{Percent Trainable} & \multicolumn{1}{c|}{\textbf{Traninable Parameters}} & \textbf{Percent Trainable} \\ \hline \hline
Fine-tuning                                           & 117M                                           & 100\%                      & 117M                                           & 100\%                      \\
CTRL                                                  & 117M                                           & 100\%                      & 117M                                           & 100\%                      \\
\hline
Prompt-tuning                                                & 0.13M                                              & 0.11\%                     & 0.21M                                              & 0.18\%                     \\
CatPrompt                                             & 0.84M                                              & 0.71\%                     & 244M                                           & 205\%                      \\
\hline
DCG (ours)                                             & 0.66M                                              & 0.56\%                     & 0.66M                                              & 0.56\%                     \\

DCG w/o DL                                      & 0.66M                                              & 0.56\%                     & 0.66M                                              & 0.56\%                     \\
\hline
\end{tabular}
}
\caption{Number of parameters used for different models. Trainable parameters is the number of parameter used for training in models. Percent Trainable is the ratio of trainable parameters to original GPT-2.}
\label{tab:trainable_parameters}
\end{table*}

\section{Inference Efficiency}
We compare the average inference efficiency of our methods with the baselines. As we can observe from Table \ref{tab:decoding_speed}, the inference speed of PPLM, FUDGE, and CoCon is far slower than the original GPT-2 model. Prompt-based methods are much faster than that decoding strategy based methods. The inference speed of our method is close to the original DialoGPT methods. As shown in Table \ref{tab:trainable_parameters}, with the growth of attribute combinations, the trainable parameters of CatPrompt increase rapidly, from 0.84M to 224M, which even exceeds the 117M trainable parameters of full DialoGPT. While our method achieves better results with a lower number of trainable parameters on DialyDialog-CG and ConvAI2-CG.

\begin{table*}[t]
\small
\begin{center}
\resizebox{0.8\textwidth}{!}{
\begin{tabular}{l|c|c|c|c|c|c|c}
\hline 
 \multirow{3}{*}{\textbf{Model}} & \multicolumn{4}{c|}{\textbf{DailyDialog-CG}} & \multicolumn{3}{c}{\textbf{ConvAI2-CG}} \\
\cline{2-8}
& \multicolumn{2}{c|}{\textbf{Controllability}}& \multicolumn{2}{c|}{\textbf{Text Quality}}& \multicolumn{1}{c|}{\textbf{Controllability}} & \multicolumn{2}{c}{\textbf{Text Quality}} \\
\cline{2-8}
& \textbf{Emo.} & \textbf{Act.} & \textbf{Flu.} & \textbf{Rel.} & \textbf{Per.} & \textbf{Flu.} & \textbf{Rel.}\\
\hline
CTRL & 2.20 & 2.05 &4.19 & 3.35&1.70 & 4.02 & 3.25  \\
DCG & 2.35 & 2.85 & 4.42& 3.89& 2.17& 4.03 & 3.26\\
DCG w/o DL & 1.70 & 2.30 & 4.04 & 3.18 & 1.61& 4.07 & 3.22 \\
\hline
\end{tabular}}
\end{center}
\vspace{-0.3cm}
\caption{Human evaluation on controllability and text quality for DailyDialog-CG and ConvAI2-CG. Emo., Act., and Per. are the attributes of emotion, act, and persona. Flu. and Rel. are the fluency and context relevancy. }
\label{tab:human_evaluation}
\end{table*}

\begin{figure*}[t]
    \centering
    \subfigure[Prompt Length]{
        \includegraphics[width=0.31\textwidth]{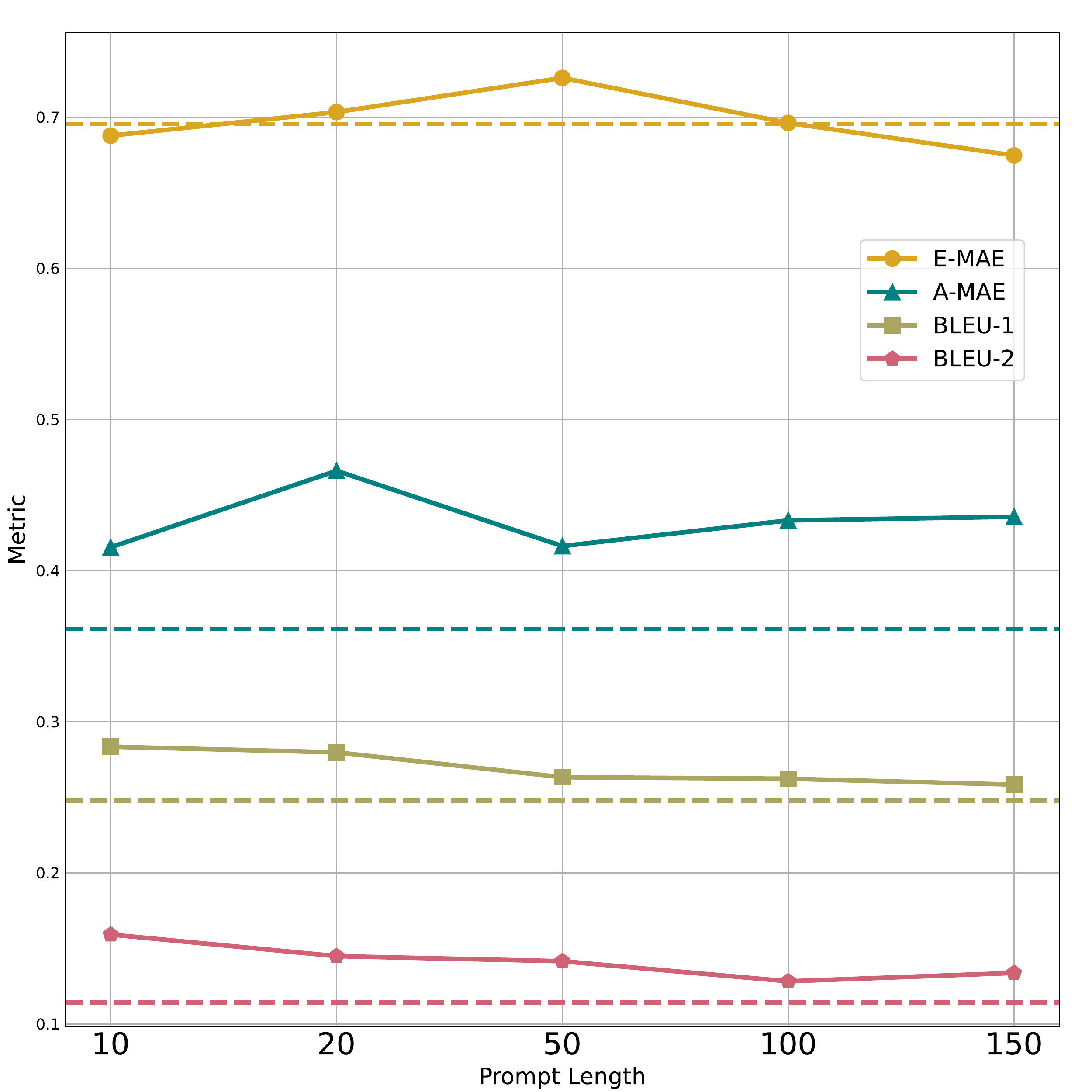}
    }
    \subfigure[Disentanglement Loss Weight]{
        \includegraphics[width=0.31\textwidth]{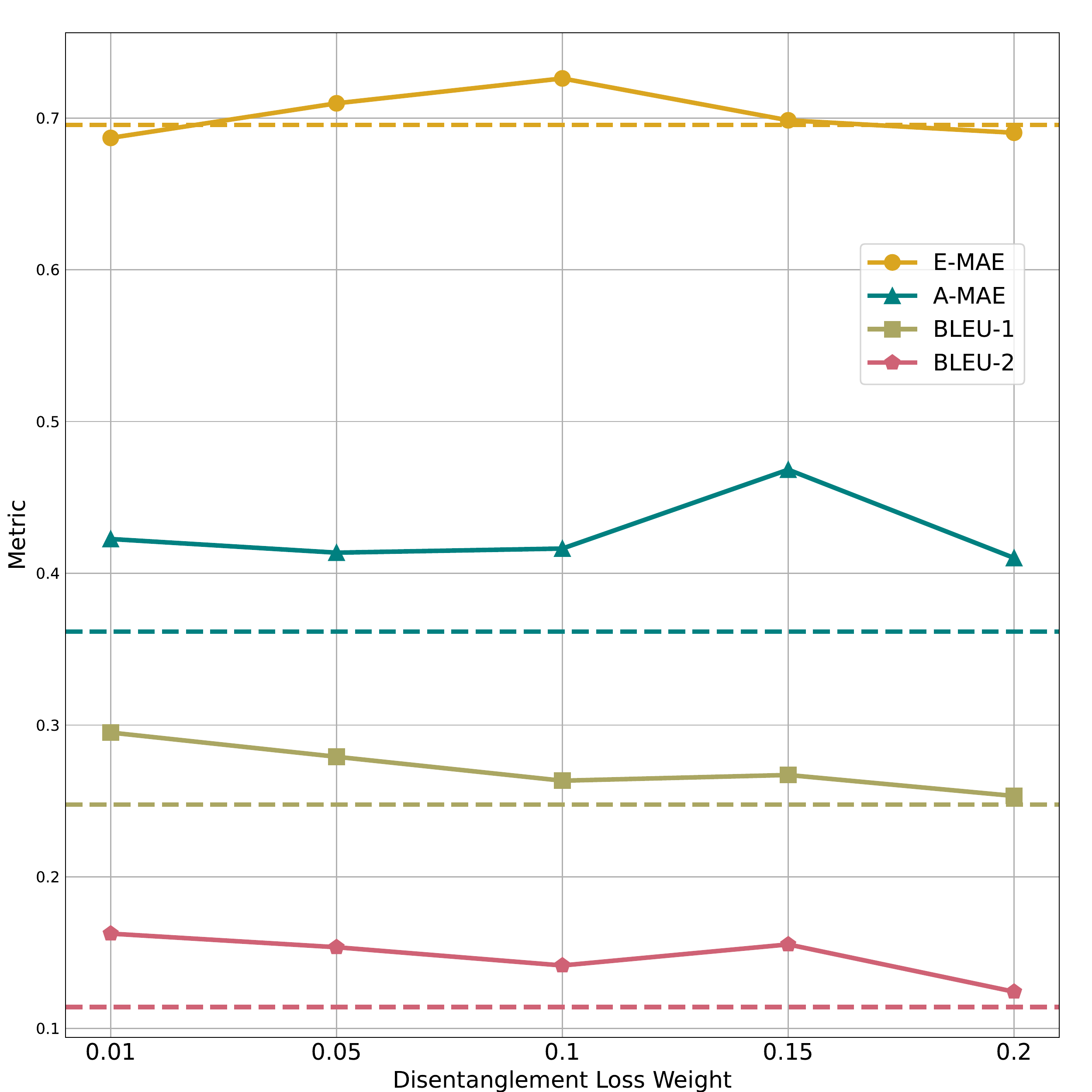}
    }
    \subfigure[Number of Pseudo Combinations]{
        \includegraphics[width=0.31\textwidth]{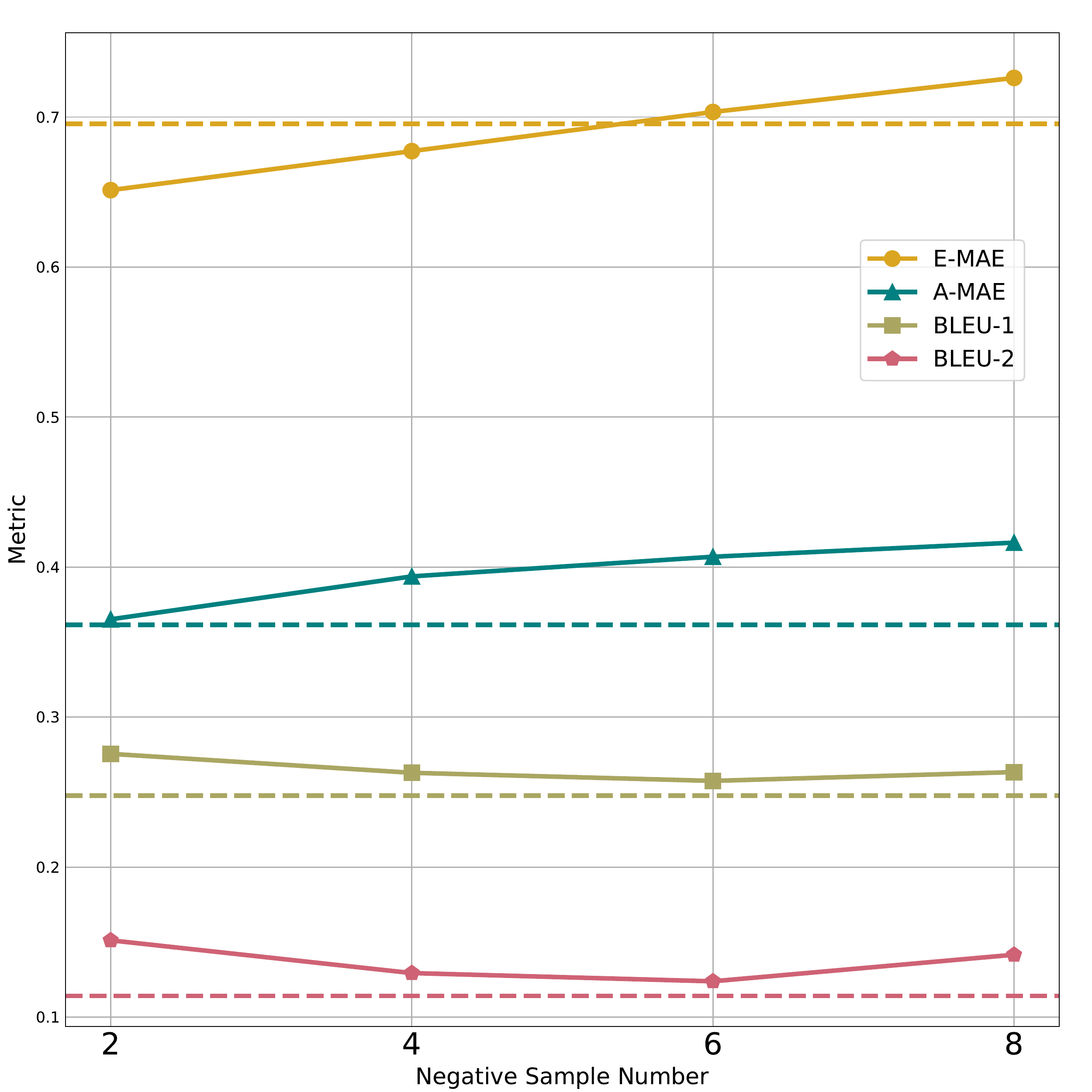}
    }
    \caption{Effect of prompt length, disentanglement loss weight, and number of pseudo combinations for DailyDialog-CG. The dotted lines denote the performance of CTRL. We report the MAE and BLEU scores for all settings.}
    \label{fig:ablation}
\end{figure*}

\section{Human Evaluation}
\label{sec:human_evaluation}
To validate the good performance of DCG, we further deploy a set of human evaluations to compare the controllability and text quality between several methods. We randomly sample 100 examples from two datasets and collect the corresponding generated responses of CTRL, DCG, and DCG w/o DL. For the controllability, 5 human annotators are invited to evaluate on a scale of 1-3, where score 1 means that the generated response is completely inconsistent with the expected attribute label, score 2 denotes that the generated response has the same meaning as the expected attribute label, but no explicit attribute-related words, and score 3 means that the generated response contains some clear attribute words. For the text quality, we ask the annotators to evaluate the fluency and context relevancy of the generated responses on a scale of 1-5, where a higher score indicates better quality. The inter-annotator agreement on the controllability and text quality is 0.63 and 0.61 for DailyDialog-GC, and 0.58 and 0.60 for ConvAI2-CG. For all metrics, the average score of the 5 annotators is treated as the final score.

As shown in Table \ref{tab:human_evaluation}, the text quality scores of all models are high,  which is because the models fine-tuned on contextualized language backbones can generate fluent sentences with relevant information. For controllability, our DCG achieves better performance than CTRL both on the coarse-grained discrete attributes and fine-grained continuous attributes, which suggests that our shared prompt mapping can learn the attribute concepts from seen attribute values to unseen attribute combinations and is useful for diverse attributes. Besides, when removing the disentanglement learning, the scores of our DCG w/o DL drop significantly, which further shows the effectiveness of the combination disentanglement  to improve the generation ability.

\begin{figure*}[t]
\centering
\resizebox{0.9\textwidth}{!}{
\includegraphics[scale=0.65]{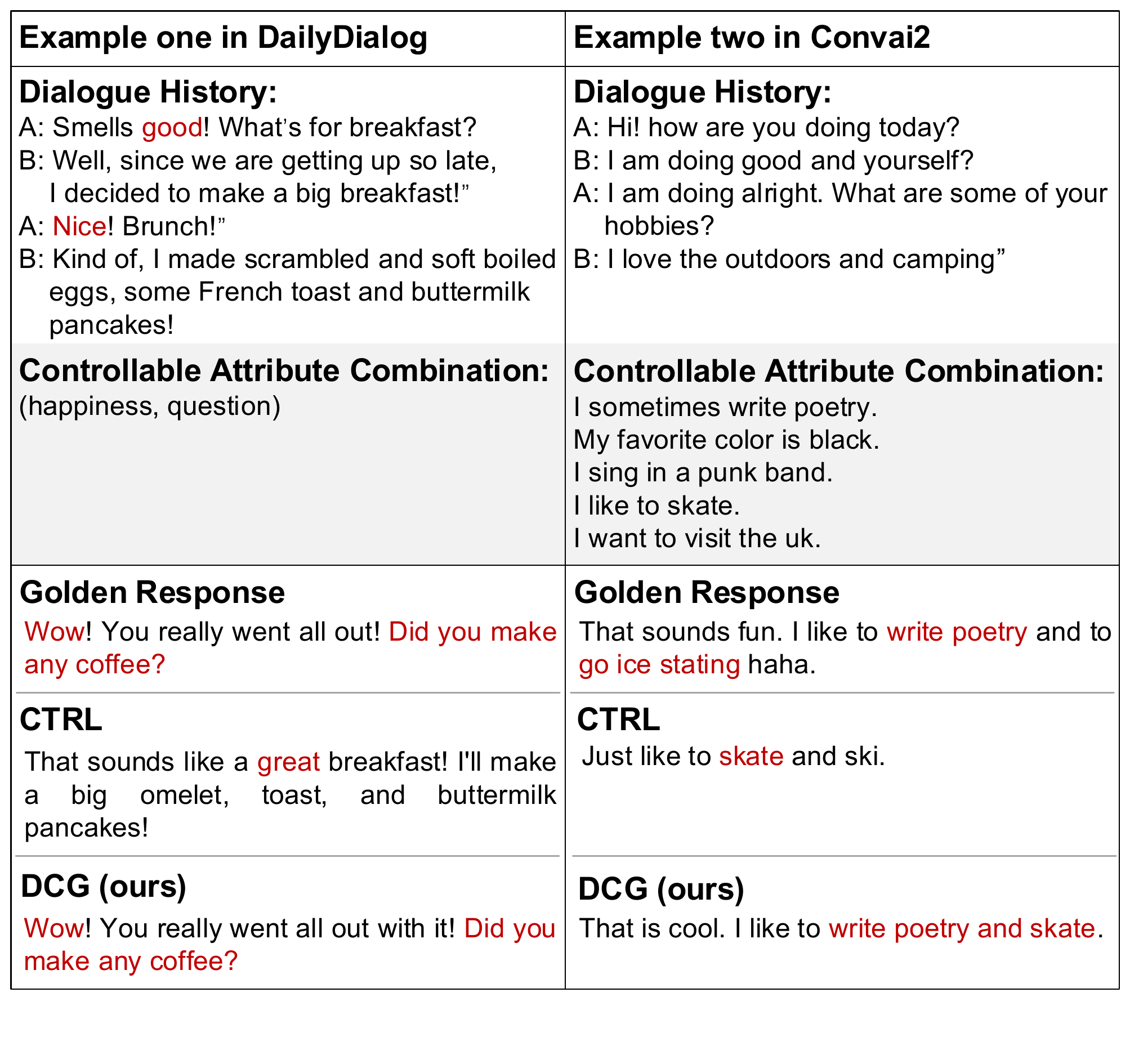}}
\vspace{-0.7cm}
\caption{Case study for two examples from DailyDialog-CG and ConvAI2-CG. We present the dialogue history, its corresponding controllable attribute combination, golden response, CTRL prediction, and prediction of our DCG.}
\label{fig:case}
\end{figure*} 

\begin{table*}[htp]
\centering
\resizebox{1.0\textwidth}{!}{
\begin{tabular}{l|ccc|ccc|ccc}
\hline
\multirow{3}{*}{\textbf{Metrics}} & \multicolumn{6}{c|}{\textbf{DailyDialog-CG}} & \multicolumn{3}{c}{\textbf{ConvAI2-CG}} \\
\cline{2-10}
     & \multicolumn{3}{c|}{\textbf{Emotion}}                           & \multicolumn{3}{c|}{\textbf{Act}}                            & \multicolumn{3}{c}{\textbf{Persona}}                            \\ \cline{2-10}
     & \textbf{Pearson}          & \textbf{Spearman}         & \textbf{Kendall}          & \textbf{Pearson}         & \textbf{Spearman}         & \textbf{Kendall}          & \textbf{Pearson}          & \textbf{Spearman}         & \textbf{Kendall}          \\ \hline
CTRLEval      & \textbf{0.6927}  & 0.6994           & 0.5961           & 0.1232  & 0.3391  & 0.2743           & 0.4059 & 0.3622
 & 0.2847 \\
MAE           & 0.6821           & \textbf{0.7500}  & \textbf{0.6242}  & \textbf{0.5446}           & \textbf{0.4661}  & \textbf{0.3936}           & \textbf{0.5793}  & \textbf{0.5768}  & \textbf{0.4418}  \\ \hline
\end{tabular}
}
\caption{Pearson ($r$), Spearman ($\rho$), and Kendall ($\tau$) correlations of attribute controllability in DailyDialog-CG and ConvAI2-CG. We use the attribute relevance of CTRLEval as the controllability score.}
\label{tab:ctrleval}
\end{table*}

\begin{table*}[t]
\small
\begin{center}
\resizebox{0.8\textwidth}{!}{
\begin{tabular}{l|ccc|ccc}
\hline
\multicolumn{1}{c|}{\multirow{2}{*}{\textbf{Model}}} & \multicolumn{3}{c|}{\textbf{Controllability}}    & \multicolumn{3}{c}{\textbf{Text Quality}}           \\ \cline{2-7} 
\multicolumn{1}{c|}{}                        & \textbf{P-SIM} $\uparrow$ & \textbf{P-NLI} $\uparrow$ & \textbf{P-MAE} $\uparrow$ & \textbf{BLEU-1} $\uparrow$ & \textbf{BLEU-2} $\uparrow$ & \textbf{METEOR} $\uparrow$ \\ \hline
CTRL                                         & 67.09          & 77.21          & 26.38          & 19.44           & 3.20            & 12.51           \\
DCG                                         & 70.63          & 83.20          & 31.18          & 18.63           & 2.32            & 11.87           \\ \hline
\end{tabular}}
\end{center}
\vspace{-0.2cm}
\caption{The performance of CTRL and DCG for ConvAI2-CG when the number of attributes varies. We train models with 4 attributes and inference with 5 attributes. Results are averaged over three random runs. ↑ means a higher score is better. (p < 0.01 under t-test)}
\label{attributes_varies}
\vspace{-0.05cm}
\end{table*}

\begin{table*}[t]
\centering
\resizebox{0.5\textwidth}{!}{
\begin{tabular}{l|ccc}
\hline
\multicolumn{1}{c|}{\textbf{Model}} & \textbf{BLEU-1 $\uparrow$} & \textbf{BLEU-2 $\uparrow$} & \textbf{METEOR $\uparrow$} \\ \hline
CTRL                                 & 24.76           & 11.42           & 20.45           \\
CTRL+TOP                             & 25.88           & \textbf{14.36}           & 21.82           \\
DCG                            & \textbf{26.33}           & 14.16           & \textbf{24.57}           \\ \hline

\end{tabular}
}
\caption{The performance of CTRL , CTRL+TOP and DCG for DailyDialog-CG. Results are averaged over three random runs. ↑ means a higher score is better. (p < 0.01 under t-test)}
\label{top_text}
\end{table*}

\section{Effect of Model Parameters}

\noindent\textbf{Prompt Length} Figure \ref{fig:ablation} (a) displays the effect of overall prompt lengths of $E_{p}$. Since the length of attribute-oriented prompt is fixed to the number of control code, we change the length of the task-oriented prompt. We find that our DCG achieves superior performance when the prompt length is between 20 and 100, and gets the best scores when the prompt length is 50. The DCG outperforms the strong baseline CTRL by the 3.19\% (averaged) for MAE and 2.16\% (averaged) for BLEUs but uses only 56\% trainable parameters of CTRL, which verifies the effectiveness and robustness of our method.

\noindent\textbf{Weight of Disentanglement Loss} Figure \ref{fig:ablation} (b) shows the effect of different weight ratios $\alpha$ for the disentanglement loss $\mathcal{L}_{D}$. We observe that $\alpha \in (0.05, 0.15)$ achieves consistent improvements than CTRL and we take $\alpha = 0.10$ in all experiments.

\noindent\textbf{Number of Pseudo Combinations} Figure \ref{fig:ablation} (c) shows the effect of the number of pseudo combinations in the disentanglement loss. We find a larger number will improve the controllability of our model. It's because more pseudo attribute values help the model to separate the desired attribute combination from the others.

\section{Comparison with CTRLEval}
Automatic evaluation metrics are important for text generation tasks, including reference-based like BLEU \cite{papineni-etal-2002-bleu}, ROUGE \cite{lin-2004-rouge}, BERTScore \cite{zhang2019bertscore} and unreferenced like perplexity \cite{brown1992estimate}, discriminator scores \cite{dathathri2019plug}, BARTScore \cite{yuan2021bartscore}. To evaluate controllability, \cite{dathathri2019plug,yang-klein-2021-fudge} trained an attribute classifier to predict the probability using labeled external data, which is hard to multi-attribute controllable generation. As a concurrent work, CTRLEval \cite{ke-etal-2022-ctrleval} proposes an evaluation method for controllable text generation. Different from our MAE, CTRLEval uses handcrafted prompts to evaluate attribute relevance.  However, handcrafted prompts are hard to construct for new tasks and cause generation bias. In contrast, our MAE uses a learnable soft prompt based on PLMs to enhance the generalization capability and robustness. We also provide a performance comparison in Table \ref{tab:ctrleval}. Results show our MAE shows superior correlations of attribute controllability.

\section{Performance on Number of Attribute}
\label{sec:Performance on Number of Attribute}
To prove our model still be useful when the number of attributes varies from training to inference, we train CTRL and our DCG with 4 attributes and inference with 5 attributes in ConvAI2-CG. As shown in Table \ref{attributes_varies}, DCG outperforms the strong baseline CTRL by 3.54\% , 5.99\%, 4.8\% in P-SIM, P-NLI and P-MAE on controllability and achieves comparable BLEU scores. It proves DCG can also handle well with changed number of attributes.

\section{Impact of TOP on Text Quality}
\label{sec: top on text qulaity}

We prove that task-oriented prompts (TOP) can also improve text quality when combined with other methods. Specifically, we trained CTRL with TOP in our experiments. As Table
\ref{top_text} shows, the results of CTRL for BLEU-1, BLEU-2, and METEOR are 24.76\%, 11.42\%, and 20.45\%, respectively. Meanwhile, the results of CTRL+TOP for BLEU-1, BLEU-2, and METEOR are 25.88\%, 14.36\%, and 21.82\%. These results indicate that CTRL can utilize TOP to enhance text quality.

\end{document}